\documentclass[11pt,a4paper]{article}

\usepackage[utf8]{inputenc}
\usepackage[T1]{fontenc}
\usepackage{graphicx}
\usepackage{amsmath}
\usepackage{amssymb}
\usepackage{booktabs}
\usepackage{colortbl}
\usepackage[margin=1in]{geometry}
\usepackage{hyperref}
\usepackage{natbib}
\usepackage{caption}
\usepackage{subcaption}
\usepackage{float}
\usepackage{algorithm}
\usepackage{algorithmic}

\title{Summer-22B: A Systematic Approach to Dataset Engineering and Training at Scale for Video Foundation Model}

\author{
Simo Ryu \\
\textit{fal} \\
\texttt{simo@fal.ai}
\and
Chunghwan Han \\
\textit{fal} \\
\texttt{hanch@fal.ai}
}

\date{October 30, 2025}

\begin{document}

\maketitle

\begin{abstract}
We describe our experience training Summer-22B, a video foundation model developed from scratch. This report documents the engineering challenges, design decisions, and lessons learned while scaling from raw footage collection to a functional model trained on approximately 50 million clips. We outline our approach combining metadata-driven dataset curation, multi-stage filtering, $\mu$P parameterization, and hypersphere-constrained optimization. We developed the Lavender Data system for dataset management and adopted inference-aware architectural choices. We share observations on what worked in our setting: dataset engineering consumed the majority of effort, architectural variants showed smaller differences than we expected, and $\mu$P hyperparameter transfer appeared effective even under geometric constraints. We hope this account proves useful to others undertaking similar projects.
\end{abstract}

\section{Introduction}
\label{sec:introduction}

The development of video foundation models represents a significant challenge in machine learning, requiring the intersection of large-scale dataset engineering, efficient training methodologies, and careful optimization strategies. Video modeling demands extensive preprocessing to transform raw footage into training-ready clips while maintaining semantic coherence and visual quality.

This work addresses the practical challenges of building video foundation models from first principles, documenting the engineering decisions and methodologies required to scale from zero initial data to a functional model. We present Summer-22B, a video diffusion model trained on approximately 50 million clips (500 billion tokens). We focus on diffusion-based architectures, which have demonstrated favorable properties in data-constrained settings compared to autoregressive approaches \citep{diffusionbeats}.

A central observation from this work is that dataset engineering and optimization methodology consumed substantially more effort than architectural exploration. Through systematic experimentation across multiple model sizes (30M to 1B parameters) and training horizons (1K to 100K steps), we find that Maximal Update Parameterization ($\mu$P) \citep{mup} enables hyperparameter transfer from small-scale experiments to larger training runs with minimal adjustment. We frame our weight-constrained optimization as \emph{Riemannian gradient descent} on the sphere manifold \citep{riemannian_optim}, maintaining weights on the hypersphere throughout training via tangent space projection and retraction. This geometric formulation eliminates the need for weight decay and its associated schedule, reducing the number of hyperparameters that must be tuned---a key practical advantage. It also constitutes a key distinction from prior work \citep{edm2} that applies normalization ad-hoc. To our knowledge, we are the first to combine $\mu$P with hypersphere-constrained Riemannian optimization, demonstrating that the two techniques are compatible and that hyperparameter transfer remains effective under geometric constraints.

Our contributions are as follows:
\begin{itemize}
\item A comprehensive video preprocessing pipeline incorporating shot boundary detection, multi-stage quality filtering, hierarchical captioning, and GPU-accelerated deduplication, scaled via Ray to process tens of millions of videos.
\item The Lavender Data system for unified dataset visualization, filtering, and streaming to training, ensuring strict parity between what engineers see and what models consume.
\item The first demonstration that $\mu$P hyperparameter transfer works under hypersphere-constrained Riemannian optimization, with empirical scaling laws for batch size ($\text{LR} \propto \sqrt{B}$) and training duration ($\text{LR} \propto 1/\sqrt{T}$).
\item Inference-aware architectural design with parallel attention-MLP computation, reducing inference latency by approximately 20\% while maintaining training stability.
\item Evaluation on VBench 1.0 \citep{huang2024vbench} and VBench 2.0 \citep{zheng2025vbench} benchmarks, providing transparent comparison against systems of similar scale. The total project cost of approximately \$300K (including \$150K compute) demonstrates the accessibility of video foundation model development.
\end{itemize}

The remainder of this paper is organized as follows. Section \ref{sec:related} surveys related work in video modeling, dataset engineering, and training methodologies. Section \ref{sec:data} details our data collection and preprocessing pipeline. Section \ref{sec:training} describes the training methodology, including architectural choices, parameterization strategies, and optimization techniques. Section \ref{sec:experiments} presents experimental results and analysis. Section \ref{sec:benchmark} evaluates the final model on VBench benchmarks. Section \ref{sec:conclusion} concludes with lessons learned and future directions.

\section{Related Work}
\label{sec:related}

\subsection{Video Foundation Models}

Video generation has advanced rapidly, driven by progress in both architectures and training methodology. Early diffusion-based approaches extended image U-Net backbones to the temporal domain \citep{ho2022video, stablediffusionvideo, animatediff}, but the field has largely converged on Diffusion Transformer (DiT) architectures \citep{peebles2023scalable} that replace convolutional backbones with pure transformer blocks. This shift enables more straightforward scaling and leverages the same training infrastructure developed for large language models.

On the proprietary side, OpenAI's Sora \citep{sora} demonstrated that DiT-based video models can produce remarkably coherent long-form video, and Meta's Movie Gen \citep{moviegen} provided one of the most detailed public accounts of large-scale video model training. The open-source ecosystem has progressed in parallel: CogVideoX \citep{cogvideox} introduced an expert transformer design, HunyuanVideo \citep{hunyuanvideo} presented a systematic training framework, Wan 2.2 \citep{wan2024} scaled open video models to 14B parameters, and Step-Video \citep{stepvideo} documented practical challenges at scale. Our work is most comparable to these open-source efforts in ambition, though we focus specifically on the interplay between dataset engineering and optimization methodology rather than architectural novelty.

Complementing generation, video understanding models such as UniVL \citep{univl} and InternVideo \citep{internvideo} have demonstrated the value of joint generative and discriminative pretraining. Vision-language foundations like CLIP \citep{clip} underpin many of these systems, providing both training signals and evaluation metrics that we also rely on.

\subsection{Dataset Engineering and Curation}

The importance of dataset quality and curation has been extensively studied in vision-language research. The original CLIP work relied on filtering web-scraped image-text pairs using query-based approaches to ensure diversity \citep{clip}. MetaCLIP \citep{metaclip} formalized this methodology, demonstrating that balancing datasets according to metadata distributions derived from caption vocabularies yields superior results compared to unbalanced collection strategies.

For video data specifically, shot boundary detection remains a fundamental preprocessing step. We employ TransNet \citep{transitionnet} for efficient and accurate detection of scene transitions using deep neural networks. For video quality assessment, we utilize DOVER \citep{dover} to provide aesthetic evaluation and filter low-quality clips.
 
The recent Cosmos technical report from NVIDIA \citep{cosmos} and work on systems like SkyReels \citep{skyreels} provide insights into industrial-scale video processing pipelines, though many implementation details remain proprietary.

\subsection{Scaling and Parameterization}

Understanding optimal scaling strategies has become central to training large neural networks efficiently. The Chinchilla work \citep{chinchilla} established compute-optimal training strategies for language models, demonstrating systematic relationships between model size, dataset size, and computational budget. However, diffusion models exhibit different scaling properties than autoregressive models, as recent work has shown \citep{diffusionbeats}.

Maximal Update Parameterization ($\mu$P) \citep{mup} addresses the challenge of hyperparameter transfer across model scales by providing principled initialization and learning rate schedules that remain optimal as model width increases. This enables reliable transfer of hyperparameters found via small-scale experiments to production-scale training runs, dramatically reducing the cost of hyperparameter search.

\subsection{Training Dynamics and Optimization}

Recent work on diffusion model training has emphasized the importance of careful parameterization and constraint strategies. EDM2 \citep{edm2} introduced hypersphere-constrained optimization for image diffusion models, demonstrating improved training stability and sample quality by constraining weight matrix rows to unit norm. This approach removes the need for explicit weight decay while providing clearer geometric interpretation of the optimization process.

Manifold-constrained optimization has a rich history in optimization literature, with Stiefel and Grassmannian manifolds representing alternative constraint spaces for neural network parameters. However, application of these techniques to large-scale transformer training remains relatively unexplored \citep{stiefel}. For distributed training at scale, systems like Megatron-LM \citep{megatron} and DeepSpeed \citep{deepspeed} provide tensor and pipeline parallelism strategies that complement data parallelism approaches like FSDP \citep{fsdp}.

% Distributed training references added: Megatron-LM and DeepSpeed

\section{Dataset Engineering}
\label{sec:data}

\subsection{Scale Requirements and Design Philosophy}

Diffusion models exhibit more favorable data efficiency compared to autoregressive models, particularly in constrained data regimes \citep{diffusionbeats}. This property derives from the iterative refinement process inherent to diffusion, which effectively increases the model's exposure to data through multiple denoising steps. For context, state-of-the-art autoregressive video models typically train on hundreds of millions to billions of clips, while recent diffusion-based systems such as Stable Video Diffusion \citep{stablediffusionvideo} have demonstrated strong results with considerably less data. Building on these observations, we find that a dataset of approximately 50 million video clips proves sufficient for training models in the billion-parameter range, substantially below the data requirements of comparable autoregressive approaches.

Treating each video clip as a sequence of spatiotemporal tokens, a typical clip contains approximately 10,000 tokens when accounting for temporal, height, and width dimensions (81 frames × 22 height patches × 40 width patches / 8 = ~9,000 tokens). This yields a total token budget of approximately 500 billion tokens for 50 million clips, which provides sufficient capacity for training models in the billion-parameter range while allowing multiple passes through the data.

A central principle guiding our approach is the avoidance of domain-specific assumptions. We impose no architectural priors specific to video structure beyond the use of three-dimensional Rotary Position Embeddings (RoPE) \citep{rope} to encode spatiotemporal positions. Similarly, we make no assumptions about the content or distribution of video data during collection. This domain-agnostic philosophy ensures generality of the learned representations.

\subsection{Metadata-Driven Collection Strategy}

Following the methodology established by CLIP \citep{clip} and formalized by MetaCLIP \citep{metaclip}, we adopt a metadata-driven collection approach based on vocabulary distributions. Rather than collecting videos randomly, we construct diverse query sets spanning various semantic categories and use these to guide footage acquisition from multiple sources. In practice, we follow a query-based footage collection strategy from heterogeneous sources, then balance the resulting corpus to match target vocabulary distributions.

This balanced sampling approach ensures representation of diverse visual concepts and prevents the dataset from being dominated by over-represented categories. The specific sources and query strategies remain proprietary, but the general methodology follows established practices in the vision-language literature.

\subsection{Video Segmentation and Shot Boundary Detection}

Raw footage typically consists of long videos spanning 10-20 minutes containing multiple distinct scenes. For training, we require shorter clips (3-30 seconds) with coherent semantic content. Learning multiple unrelated contexts within a single training example would complicate the optimization process and reduce data efficiency.

We distinguish between acceptable and unacceptable transitions within clips. Acceptable transitions include gradual scene evolution such as time-lapses or smooth camera motion, where semantic content remains coherent throughout. Unacceptable transitions include abrupt cuts, rapid scene changes, or pans between semantically unrelated content. These transitions would force the model to learn discontinuous jumps rather than smooth video dynamics.

We use a two-stage procedure that first applies PySceneDetect \citep{pyscenedetect} (fast, heuristic-based) to obtain initial scene splits, then compensates for its false negatives with TransNetV2 \citep{transitionnet} for fine-grained boundary detection. This split-with-cheap-then-compensate-with-accurate strategy maintains high throughput while recovering missed transitions. This approach is computationally efficient because the total frame count processed by the more expensive TransNetV2 model is significantly reduced after the initial coarse segmentation.

\subsection{Engineering Challenges at Scale}

Processing tens of millions of videos presents significant computational and engineering challenges. The trade-off between CPU and GPU utilization requires careful consideration, as GPUs provide superior throughput for video decoding and encoding but incur substantially higher costs, particularly in cloud environments.

Decoding was consistently the primary bottleneck; we benchmarked multiple decoding libraries across resolutions/codecs and selected PyNvVideoCodec \cite{pynvvideocodec} for most workloads due to strong throughput. To reach stability at high throughput, we introduced explicit methods to lock/unlock the frame buffer to avoid race-condition-driven corruption, which was a major issue with PyNvVideoCodec. We pipeline decoding in chunks and pass frames immediately to downstream stages, maximizing overlap between CPU and GPU jobs throughout the system.

\begin{figure}[H]
\centering
\includegraphics[width=0.48\textwidth]{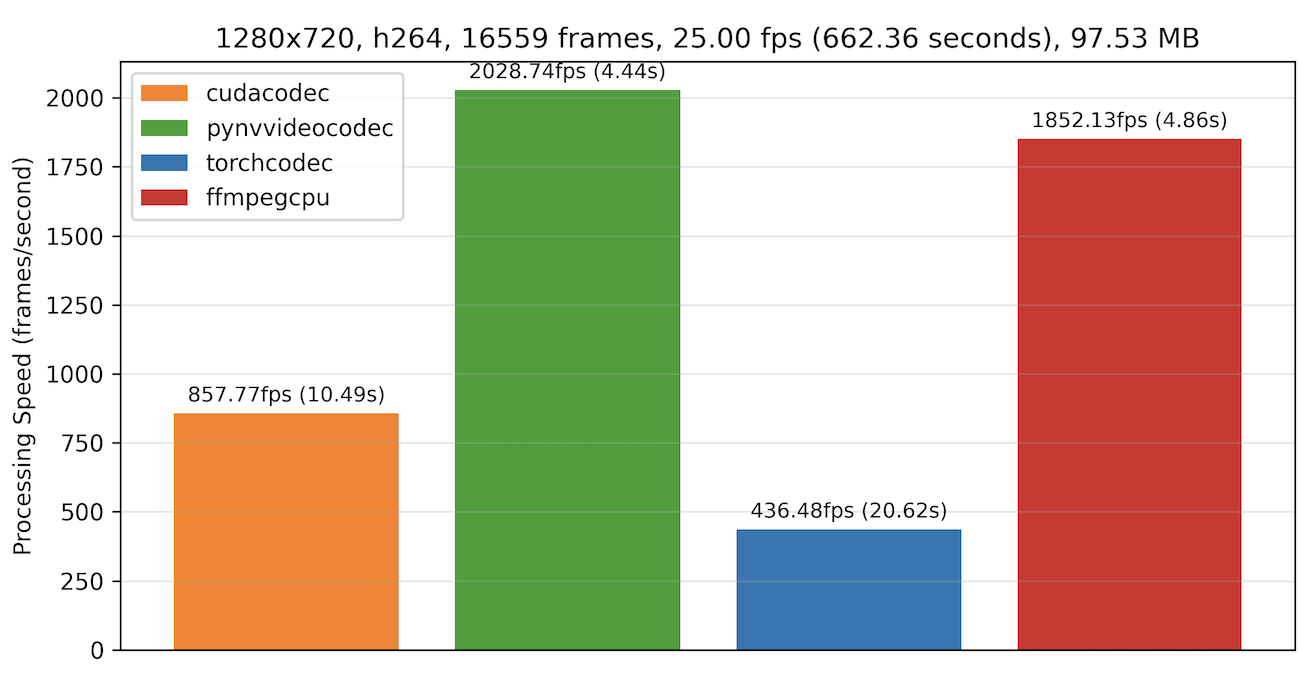}
\includegraphics[width=0.48\textwidth]{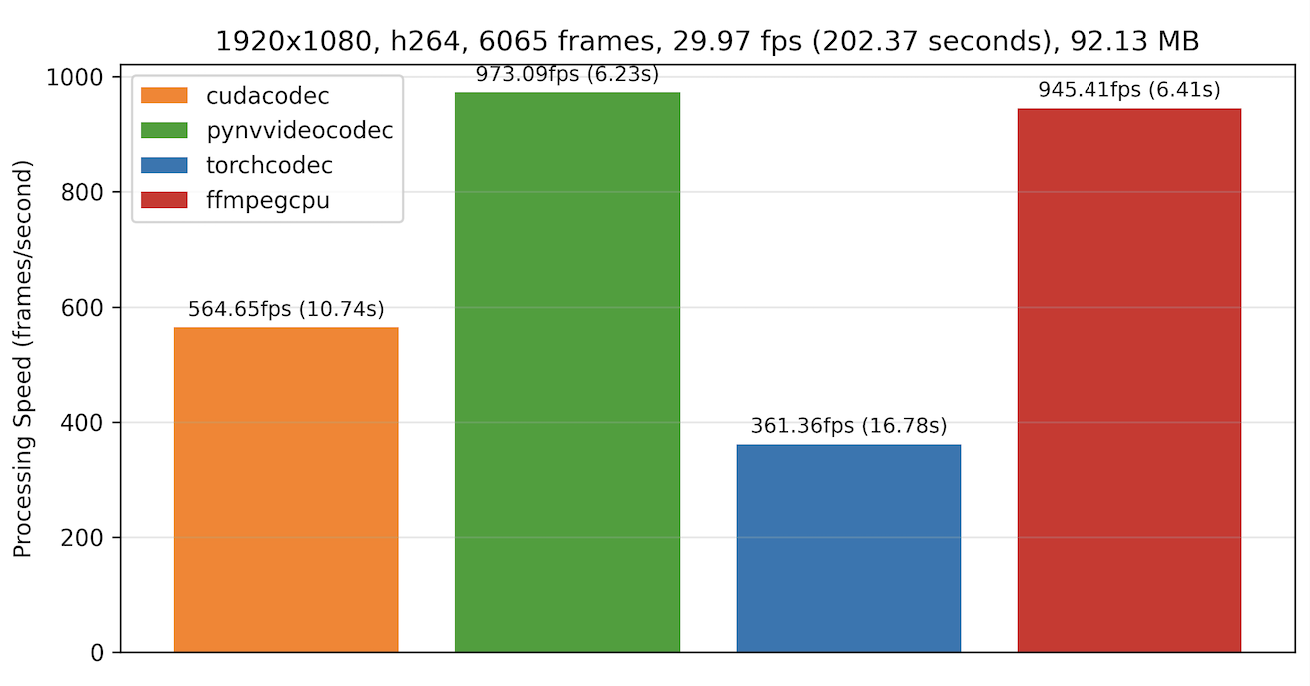}
\includegraphics[width=0.48\textwidth]{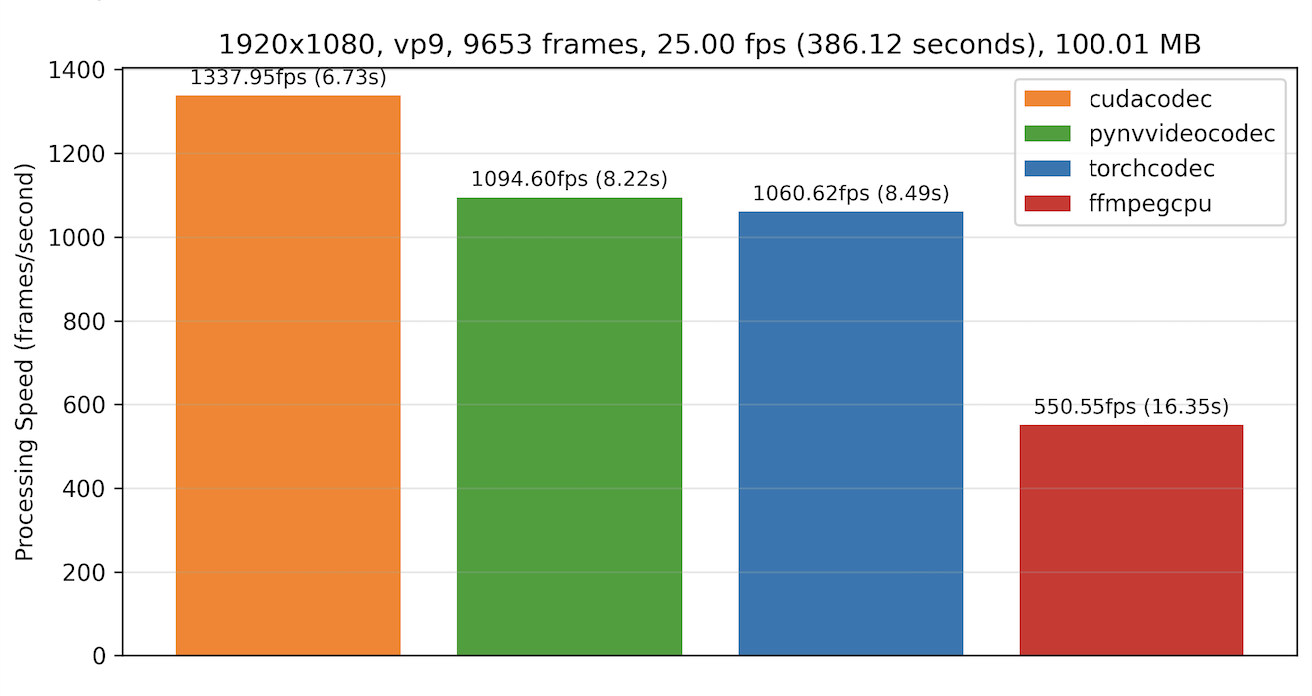}
\includegraphics[width=0.48\textwidth]{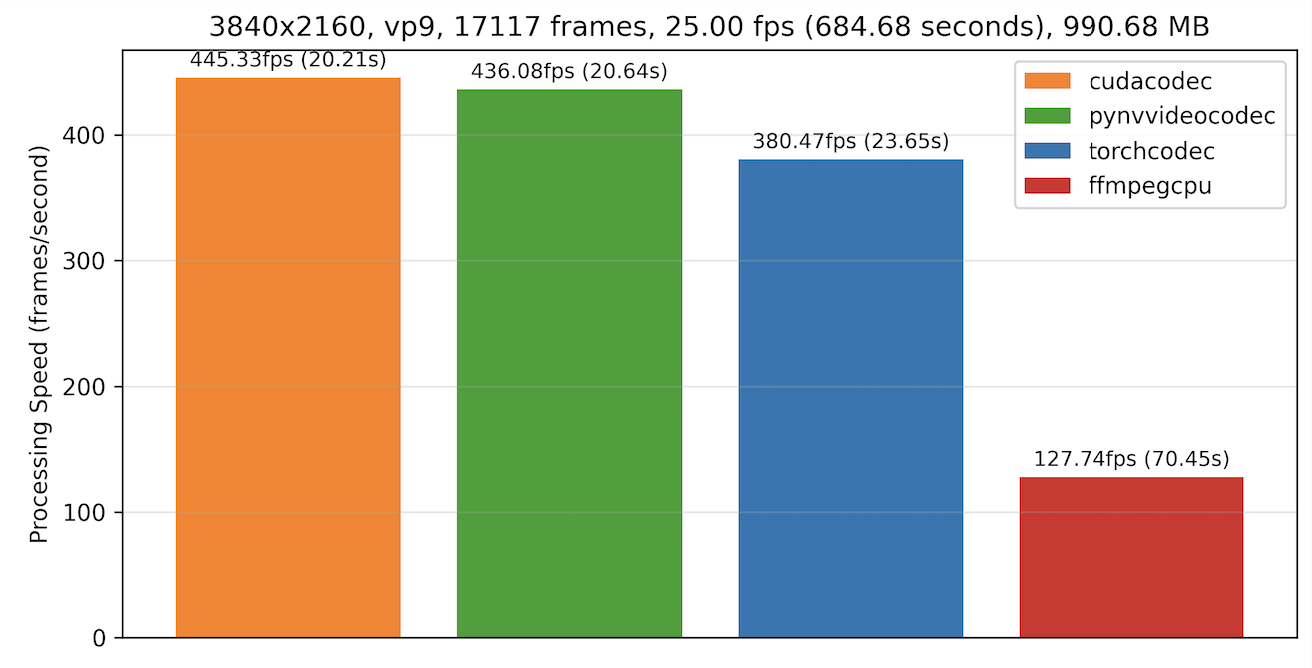}
\label{fig:decoder_benchmark}
\caption{Benchmark results of video decoding libraries across resolutions and codecs (720p H.264, 1080p H.264, 1080p VP9, 2160p VP9). PyNvVideoCodec consistently achieves the highest throughput and was selected as the default decoder in our pipeline.}
\end{figure}

To achieve efficient GPU utilization, we developed a critical optimization involving zero-copy transfer of video data between OpenCV CUDA datatypes and PyTorch tensors to avoid expensive memory transfers. We constructed desired data types from the GPU pointer in C++ level and wrapped them as PyTorch C++ extensions. This was crucial due to the fact that each library has its own preferred type (torch.Tensor vs. cv2.GpuMat) and we need to convert between them without copy.

\begin{table}[H]
\centering
\caption{GPU matrix types used across libraries in our video preprocessing pipeline. Different libraries operate on different CUDA tensor representations; zero-copy conversion between these types is critical for avoiding costly memory transfers during decoding, encoding, and inference stages.}
\begin{tabular}{lll}
\toprule
Library & Task & CUDA type \\
\midrule
PyNvVideoCodec & decoding & \texttt{torch.Tensor} \\
OpenCV \texttt{cv2.cudacodec} & encoding & \texttt{cv2.GpuMat} \\
RapidOCR & OCR & \texttt{torch.Tensor} \\
OpenCV Farneb\"ack & Flow estimation & \texttt{cv2.GpuMat} \\
DOVER (ONNX) & Aesthetic scoring & \texttt{torch.Tensor} \\
\bottomrule
\end{tabular}
\end{table}

We convert between these without copy by constructing \texttt{cv2.GpuMat} views from CUDA memory (e.g., \texttt{cv2::cuda::createGpuMatFromCudaMemory}) and by creating Torch tensors from raw device pointers (e.g., \texttt{torch::from\_blob}) where safe.

The asynchronous pipeline architecture distributes different processing stages across available resources based on their computational characteristics. Network-intensive operations (downloading and uploading) are separated from CPU-intensive tasks (shot boundary detection) and GPU-intensive operations (decoding, aesthetic scoring). Some tasks are resource-interchangeable (CPU $\leftrightarrow$ GPU), and some can operate on downscaled frames (e.g., optical flow). When possible, we prefer CPU with downscaled inputs to free GPU time for decoding/encoding. This pipelining strategy overlaps CPU and GPU jobs to minimize bottlenecks and maintain high throughput across the entire system.

\begin{table}[H]
\centering
\caption{Qualitative resource intensity by task in the preprocessing pipeline. Each stage has a dominant resource bottleneck (network, CPU, or GPU). We exploit this heterogeneity by pipelining stages so that CPU-bound and GPU-bound work overlap, maximizing overall throughput.}
\begin{tabular}{lccc}
\toprule
\textbf{Task} & \textbf{Network} & \textbf{CPU} & \textbf{GPU} \\
\midrule
Footage download & Heavy & Light & -- \\
Decoding & -- & Medium & Heavy \\
Shot detection (PySceneDetect) & -- & Heavy & -- \\
Aesthetic scoring & -- & Light & Heavy \\
Optical flow & -- & Heavy & -- \\
Encoding & -- & Medium & Heavy \\
Uploading clips & Heavy & Light & -- \\
\bottomrule
\end{tabular}
\end{table}

A persistent challenge in large-scale data processing involves long-tail errors that manifest only after processing substantial portions of the dataset. These "unknown unknowns" cannot be predicted before encountering them, and their frequency typically follows a power-law distribution (informally, the $n$-th unique bug occurs at $\mathcal{O}(1/n)$ frequency as the dataset size grows). Consequently, multiple complete passes through the dataset may be required to identify and address rare failure modes such as specific codec corruptions or metadata anomalies. Robust logging of as many parameters as possible and acquiring resume-ability rather than trying to predict the bugs beforehand greatly improves iteration speed.

\subsection{Scaling with Ray}

To scale the end-to-end preprocessing pipeline across multiple nodes, we encapsulate downloading, processing, and uploading into Ray \citep{ray} actors. Network communication is overlapped with on-node processing, and within the processing actor we further overlap CPU-bound and GPU-bound stages via threads to maximize utilization. A lightweight task distributor fans out work to many actors across the cluster.

In a representative large run, we utilized approximately $\sim$100 hours on 20 nodes with 10 L40S GPUs per node. This setup sustained high throughput by: (1) streaming inputs to actors while prior batches were still being encoded/uploaded, (2) overlapping GPU decoding/encoding with CPU-side analysis (e.g., shot detection), and (3) minimizing GPU memory transfers through zero-copy paths.

\begin{figure}[H]
\centering
\includegraphics[width=0.9\textwidth]{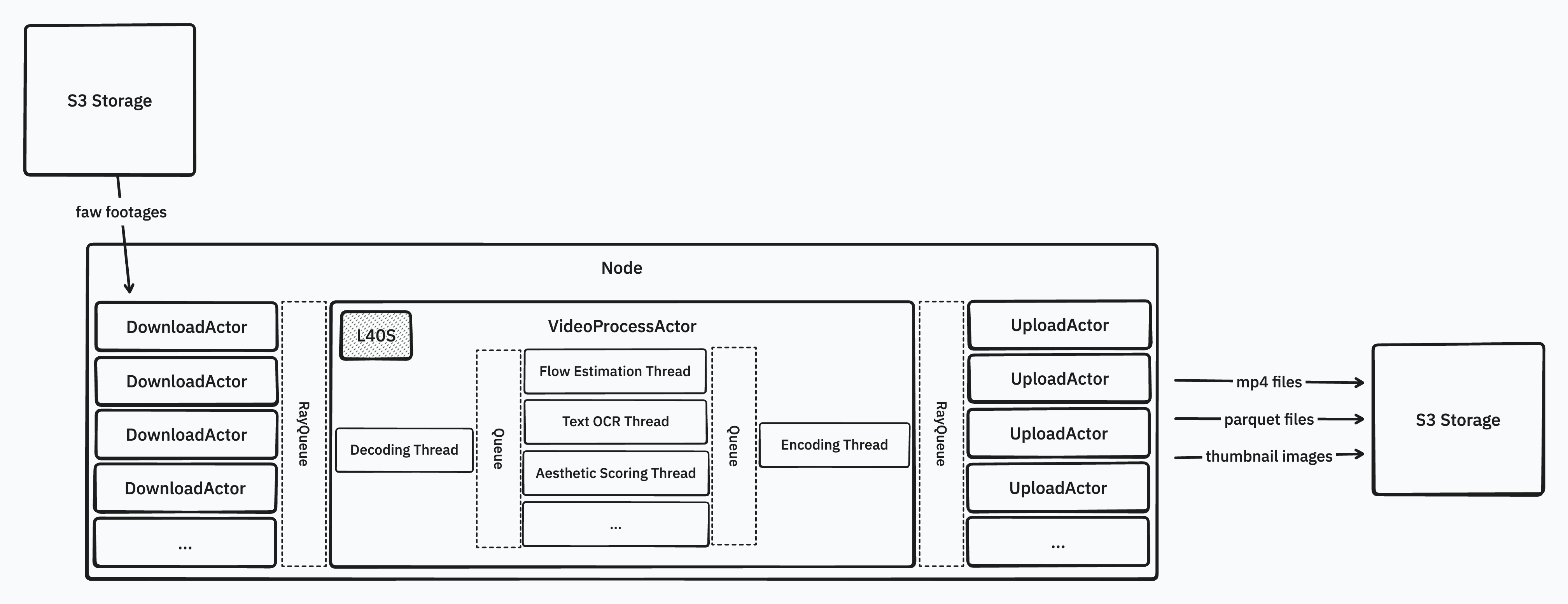}
\caption{Within-node Ray actor design: network I/O overlapped with processing; CPU and GPU jobs overlapped via threads inside the actor.}
\label{fig:ray_actor}
\end{figure}

\begin{figure}[H]
\centering
\includegraphics[width=0.9\textwidth]{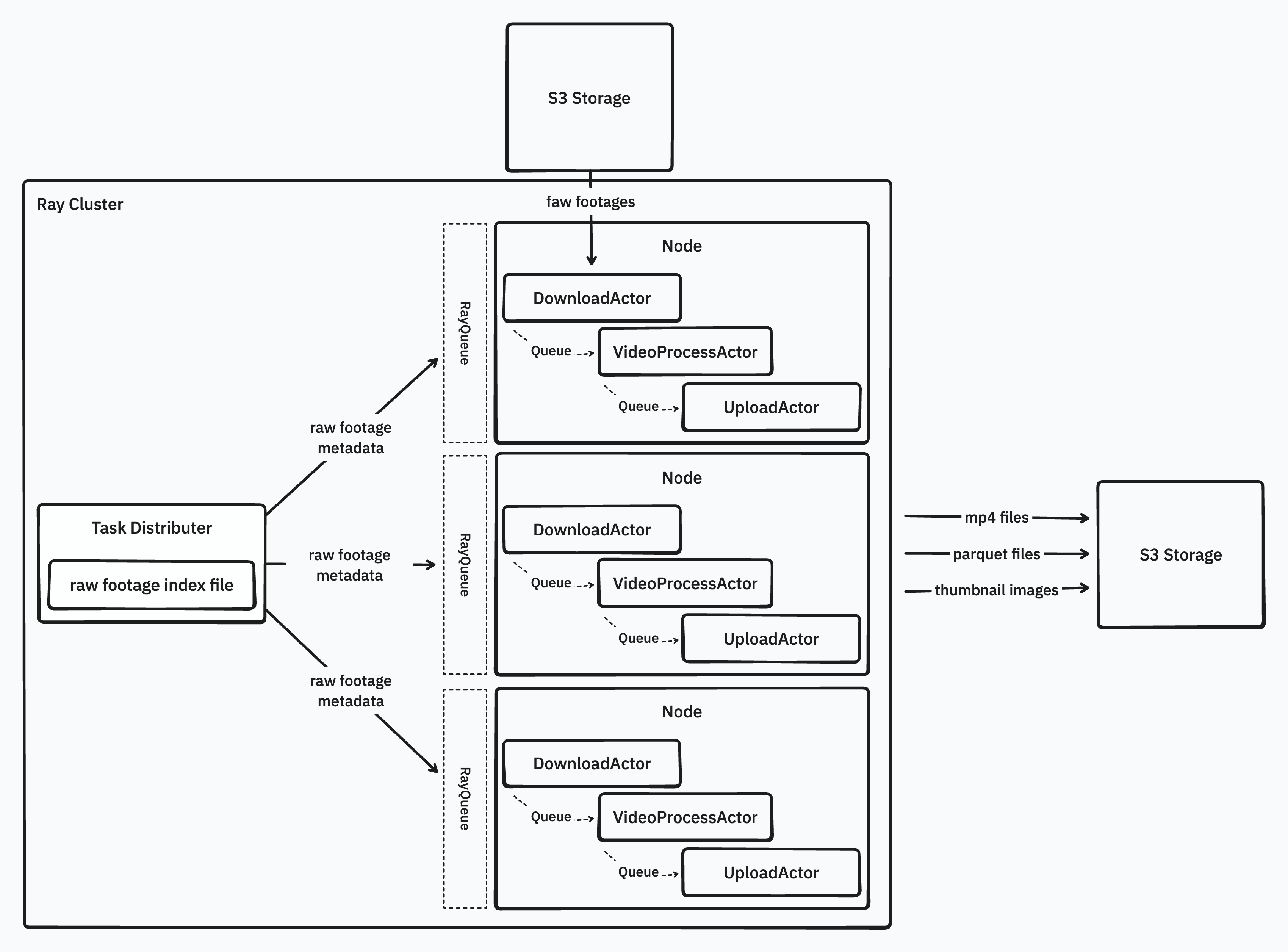}
\caption{Cluster scaling via a task distributor front-end dispatching work to many Ray actors across nodes.}
\label{fig:ray_cluster}
\end{figure}

\subsection{Multi-Stage Filtering Pipeline}

\begin{figure}[H]
\centering
\includegraphics[width=\textwidth]{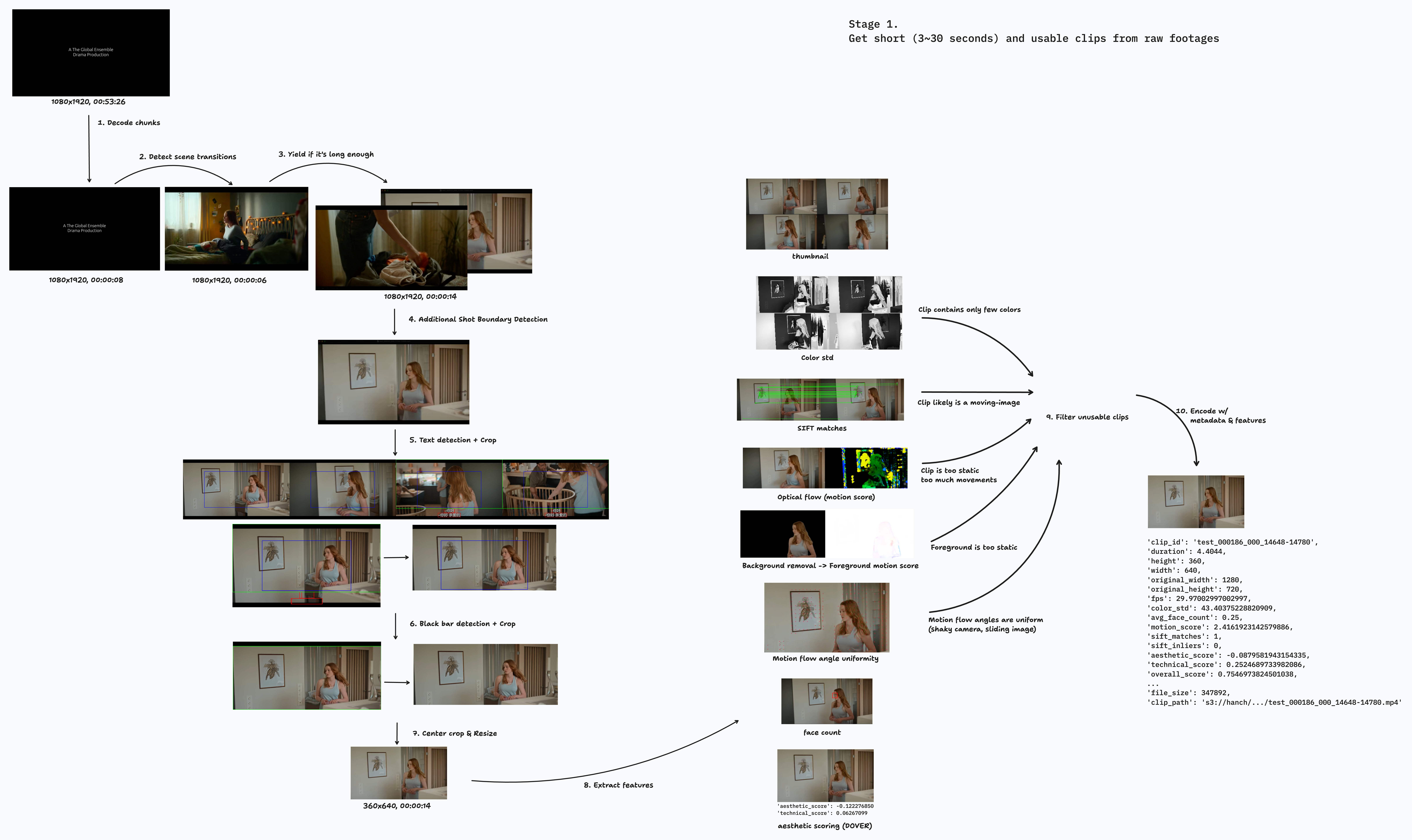}
\caption{Complete video preprocessing pipeline showing the flow from raw footage through segmentation, filtering, and encoding stages. Multiple parallel filters operate on different quality dimensions to ensure only high-quality clips proceed to training.}
\label{fig:pipeline}
\end{figure}

Figure \ref{fig:pipeline} illustrates our complete preprocessing pipeline from raw footage to training-ready clips. The pipeline consists of several stages, each addressing different quality criteria.

Following initial segmentation via shot boundary detection, we apply length constraints to ensure clips fall within the target duration range of 3-30 seconds. Text and black bar detection enables cropping to active content regions and standardizing aspect ratios. All clips are resized to a standard resolution of 352×640 pixels (maintaining 16:9 aspect ratio) with at least 81 frames at 24 FPS, corresponding to at least 3.375 seconds per clip.

We employ a multi-stage filtering pipeline to curate the dataset, where each component is designed to target specific failure modes. The collective application of these filters ensures that only high-quality, dynamic, and visually engaging clips are selected for the training set.

\subsubsection{Visual Filters}

Visual quality filters assess color diversity to remove monochrome or heavily tinted clips, detect static or near-static content via thumbnail analysis to reject slideshows and title screens, and identify potential duplicates through perceptual or SIFT \citep{sift} matches.

\subsubsection{Motion Filters}

Motion analysis incorporates optical flow estimation to quantify both overall motion and the distinction between foreground and background dynamics. We used BirefNet \citep{birefnet} for separation of foreground and background motion, and Farneb\"ack algorithm \citep{farneback} for flow estimation.

In addition to aggregate motion magnitude, we characterize video dynamics using two complementary metrics: spatial uniformity and temporal consistency. The former measures the similarity of motion across pixels within a single frame, while the latter assesses the stability of motion direction across the sequence. Table \ref{tab:motion_metrics} illustrates how different combinations of these metrics correspond to various video characteristics and their perceived quality for training.

\begin{table}[H]
\centering
\caption{Classification of video dynamics based on optical flow angle consistency (temporal stability of motion direction) and spatial uniformity (similarity of motion across pixels). The combination of these two metrics discriminates between desirable training content (e.g., parallax and tracking shots) and undesirable artifacts (e.g., sliding images, shaky camera).}
\label{tab:motion_metrics}
\begin{tabular}{cccl}
\toprule
\textbf{Consistency} & \textbf{Uniformity} & \textbf{Preferred} & \textbf{Characteristics} \\
\midrule
High & High & Bad & Sliding images, static 2D pans \\
High & Low  & Good & Complex camera motion, parallax, tracking shots \\
Low  & High & Bad & Shaky camera, sudden jumps \\
Low  & Low  & Neutral & Complex patterns (rain, snow, anime motion) \\
\bottomrule
\end{tabular}
\end{table}

\subsubsection{Content Filters}

Face counting tracks human presence as a proxy for content diversity, helping to balance the dataset against the "talking heads" over-representation described later.

The final quality gate employs DOVER \citep{dover}, a learned aesthetic scoring model that provides neural network-based quality assessment. Only clips passing all filter thresholds proceed to the encoding stage, where they are compressed to efficient formats suitable for distributed training.

\subsection{Hierarchical Captioning and Deduplication}

A significant challenge in video datasets is the overrepresentation of certain content types, particularly videos featuring talking heads. This phenomenon has been widely reported in video modeling literature and requires explicit mitigation through data balancing.

We employ hierarchical captioning at three levels of granularity: detailed captions providing comprehensive scene descriptions, short captions highlighting key elements, and ultra-short three-word captions serving as semantic buckets. This approach follows the spirit of MetaCLIP's vocabulary-based balancing but adapted for video content.

\begin{figure}[H]
\centering
\includegraphics[width=0.6\textwidth]{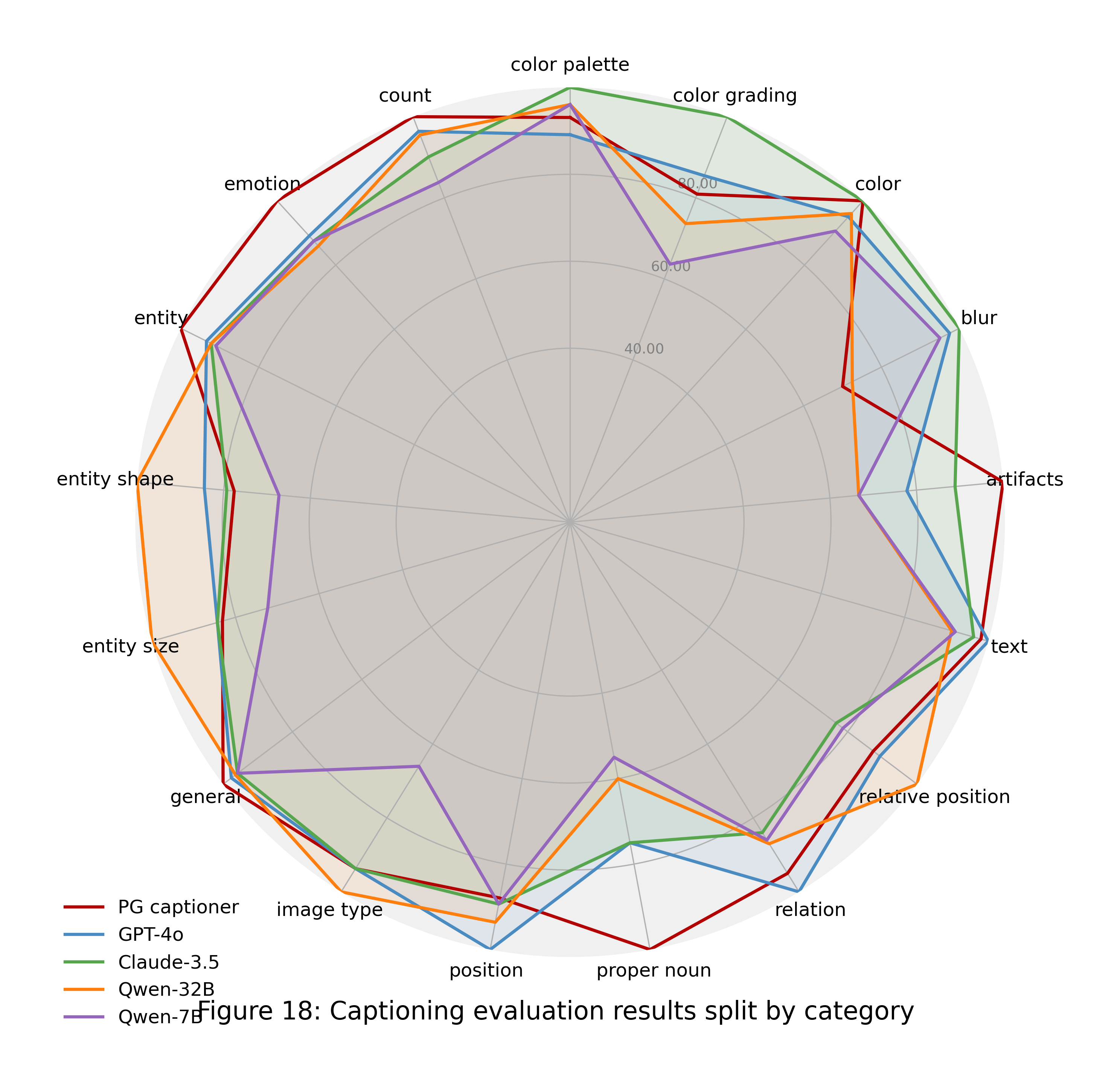}
\caption{Performance profile of the fine-tuned Qwen 2.5 VL captioning model showing inference throughput across different batch sizes and model configurations.}
\label{fig:captioner}
\end{figure}

For caption generation, we fine-tune Qwen 2.5 VL \citep{qwen2vl} on our specific captioning task, optimizing the trade-off between caption quality and inference speed. During training, we apply dynamic balancing to address gender representation: we identify clips with man or woman subjects through regex matching on very short captions, then subsample to maintain a 1:1 gender ratio capped at 20 percent of the total dataset. This prevents over-representation of any single demographic category. Figure \ref{fig:captioner} shows the performance characteristics of our captioning model.

\begin{figure}[H]
\centering
\includegraphics[width=0.6\textwidth]{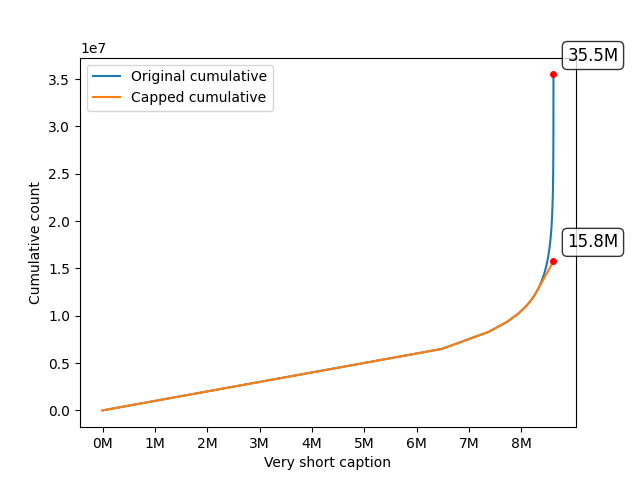}
\caption{Distribution of ultra-short three-word captions across the dataset, showing the long-tail distribution with approximately 80\% of clips having unique captions.}
\label{fig:dedup}
\end{figure}

The ultra-short three-word captions serve as semantic buckets for deduplication. Empirically, we observe that approximately 80\% of clips receive unique three-word captions, indicating the long-tail nature of video content distribution (Figure \ref{fig:dedup}). Within each bucket defined by the ultra-short caption, we perform embedding-based deduplication to remove near-duplicates while preserving diversity.

\subsection{GPU-Accelerated Embedding Clustering}

Within each semantic bucket defined by ultra-short captions, we perform embedding-based clustering to identify and remove near-duplicates. At our scale (tens of millions of clips with high-dimensional embeddings), standard clustering implementations become prohibitively slow. We developed a GPU-accelerated Mini-Batch K-means implementation with several enhancements for robustness and efficiency.

The core challenge with standard K-means initialization (K-means++) is that it requires many sequential distance computations that cannot be parallelized efficiently on GPUs. We instead use Bradley-Fayyad refinement \citep{bradleyrefining}, which works as follows: take multiple small random samples of the data, cluster each sample independently, then pool all the resulting centroids and cluster them again to pick the final starting points. This finds good initial centroids while keeping all the heavy computation parallel.

During training, we process random mini-batches rather than the full dataset. Each centroid moves toward the average of its assigned points, but we weight the update by how many points the cluster has seen historically: new clusters move quickly while established clusters are more stable. Periodically, we check for two pathological cases: clusters that have attracted almost no points (dead clusters) get moved to regions far from any existing centroid, and clusters that have grown much larger than average get split by seeding a new centroid from a random point within them. These heuristics maintain balanced cluster sizes without manual tuning.

Algorithm~\ref{alg:clustering} presents the complete procedure.

\begin{algorithm}[H]
\caption{GPU-Accelerated Mini-Batch K-means with Bradley-Fayyad Initialization}
\label{alg:clustering}
\begin{algorithmic}[1]
\REQUIRE Dataset $X$ of $N$ embeddings, number of clusters $K$, subsamples $J$
\STATE \textbf{// Bradley-Fayyad Initialization}
\STATE $C_{\text{seed}} \leftarrow$ K-means++ on $X$ \COMMENT{crude initial centroids}
\FOR{$j = 1$ to $J$}
    \STATE $S_j \leftarrow$ random 1\% subsample of $X$
    \STATE $C_j \leftarrow$ K-means($S_j$, init=$C_{\text{seed}}$, iters=20)
\ENDFOR
\STATE $C_{\text{pool}} \leftarrow$ concatenate all $C_j$ \COMMENT{$J \times K$ candidate centroids}
\STATE $C \leftarrow$ best K-means($C_{\text{pool}}$, $K$) over $J$ random restarts
\STATE
\STATE \textbf{// Mini-Batch Updates}
\STATE Initialize counts $n_k \leftarrow 0$ for each cluster $k$
\FOR{each iteration}
    \STATE $B \leftarrow$ random mini-batch from $X$
    \STATE Assign each $x \in B$ to nearest centroid
    \FOR{each cluster $k$ with assignments in $B$}
        \STATE $\eta \leftarrow n_k^{\text{batch}} / (n_k + n_k^{\text{batch}})$ \COMMENT{adaptive learning rate}
        \STATE $c_k \leftarrow c_k + \eta \cdot (\bar{x}_k^{\text{batch}} - c_k)$
        \STATE $n_k \leftarrow n_k + n_k^{\text{batch}}$
    \ENDFOR
    \STATE
    \STATE \textbf{// Cluster Maintenance (every 10 iterations)}
    \FOR{each cluster $k$ where $n_k < 0.01 \cdot \max(n)$}
        \STATE $c_k \leftarrow$ point in $B$ furthest from all centroids \COMMENT{reseed dead}
    \ENDFOR
    \FOR{each cluster $k$ where $n_k > 12 \cdot \text{mean}(n)$}
        \STATE Split $k$: seed new centroid from random point in cluster $k$
    \ENDFOR
\ENDFOR
\RETURN Centroids $C$, assignments
\end{algorithmic}
\end{algorithm}

All distance computations use batched matrix operations on GPU, enabling the implementation to process 100K points with 50 clusters in under 2 seconds (compared to several minutes for scikit-learn's CPU implementation), allowing interactive iteration on clustering parameters during dataset development.

\subsection{The Lavender Data System}

The complexity of managing tens of millions of video clips across heterogeneous storage and compute environments necessitated the development of a specialized data management infrastructure, which we designate the Lavender Datas System. Lavender Data is designed to unify the three pillars of dataset engineering: visualization, filtering, and loading.

A primary technical requirement for Lavender Data is ensuring strict parity between the data displayed in the visualization interface and the data consumed by the model during training. This consistency allows for rapid iteration on filtering thresholds and quality metrics, as engineers can immediately observe the impact of filtering decisions on the training distribution.

The system facilitates dataset evolution through a stream-merging architecture, where new features (typically stored as sharded Parquet files) can be integrated into the primary dataset without requiring expensive re-writing.

To sustain the throughput requirements of large-scale training, Lavender Data adopts a pipeline parallelism approach to data loading. Computationally intensive pre-processing, such as text encoding and VAE encoding, is distributed across dedicated nodes. These nodes stream processed tensors and attributes to the training cluster via a high-performance network interface, enabling full overlap between data preparation and gradient computation. To ensure robustness during long-horizon training runs, the system incorporates comprehensive fault-tolerance mechanisms, including the ability to skip corrupted samples or recover from transient network failures without stalling the optimization process. This architecture enables managing the entire data-to-training lifecycle with minimal manual intervention.

\begin{figure}[H]
\centering
\includegraphics[width=0.8\textwidth]{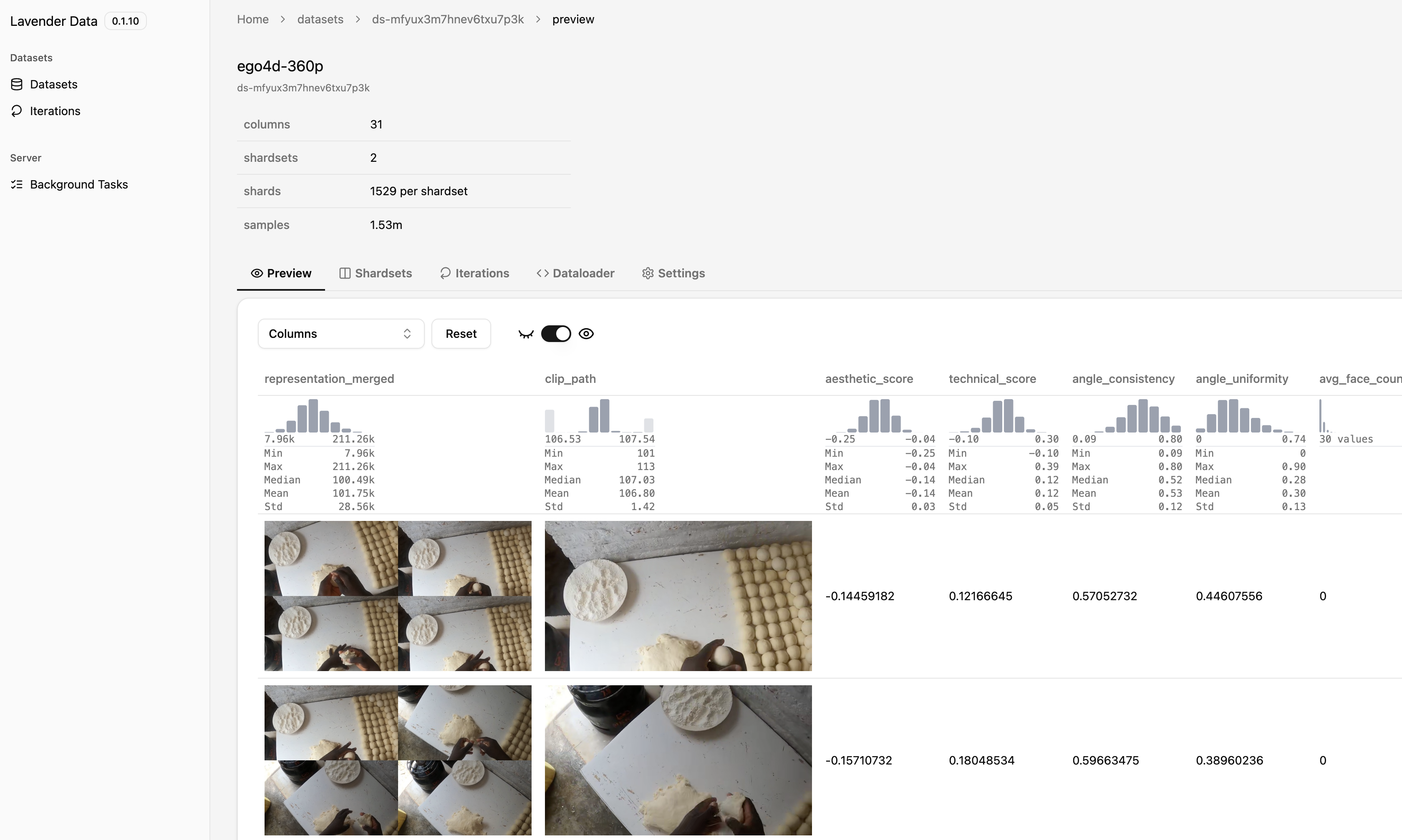}
\caption{Lavender Data web interface showing dataset visualization.}
\label{fig:lavender-web}
\end{figure}

\begin{figure}[H]
\centering
\includegraphics[width=0.8\textwidth]{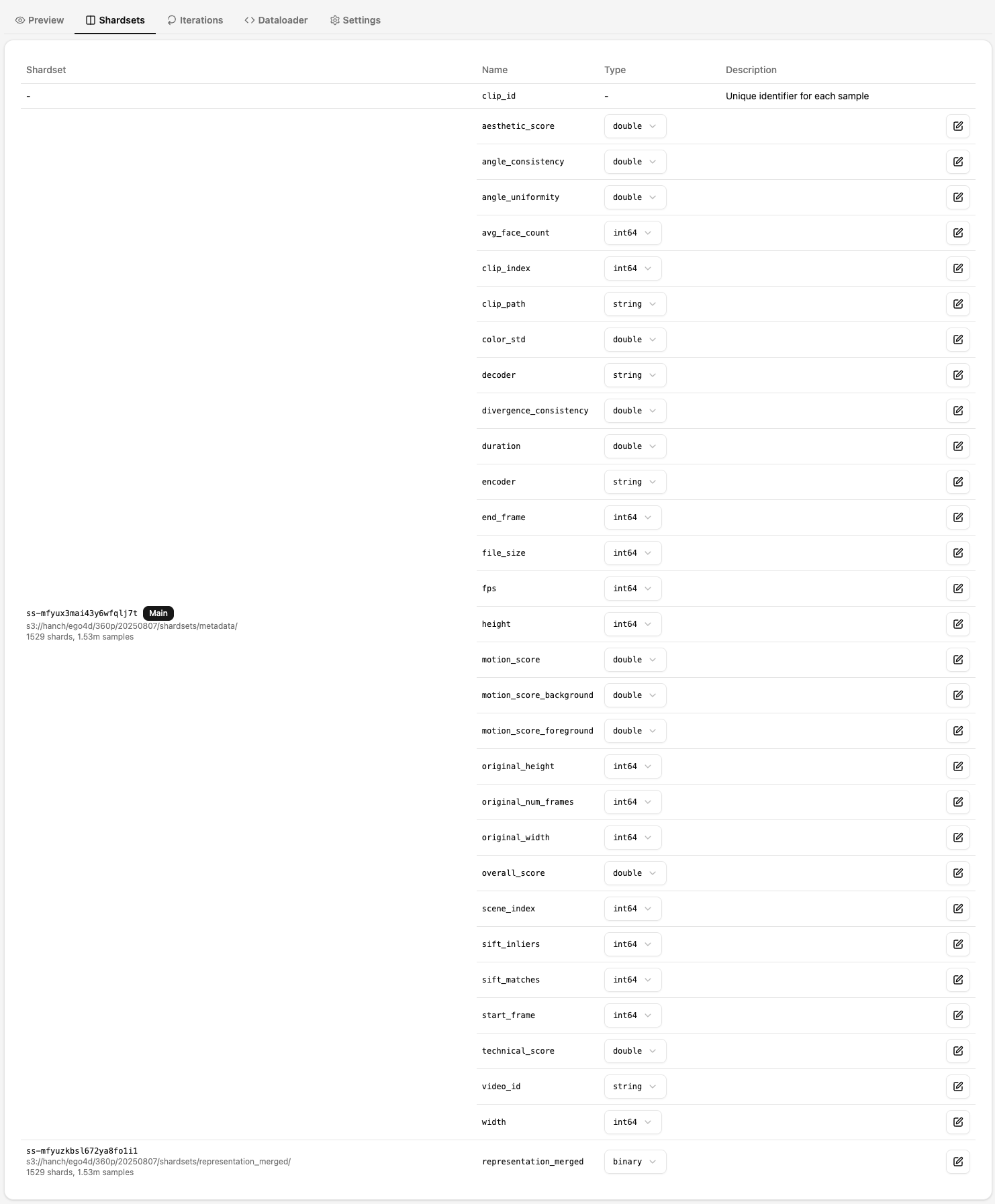}
\caption{Lavender Data web interface showing stream-merging architecture. New features can be added to the dataset without requiring re-writing existing dataset.}
\label{fig:lavender-shardset}
\end{figure}

\begin{figure}[H]
\centering
\includegraphics[width=0.8\textwidth]{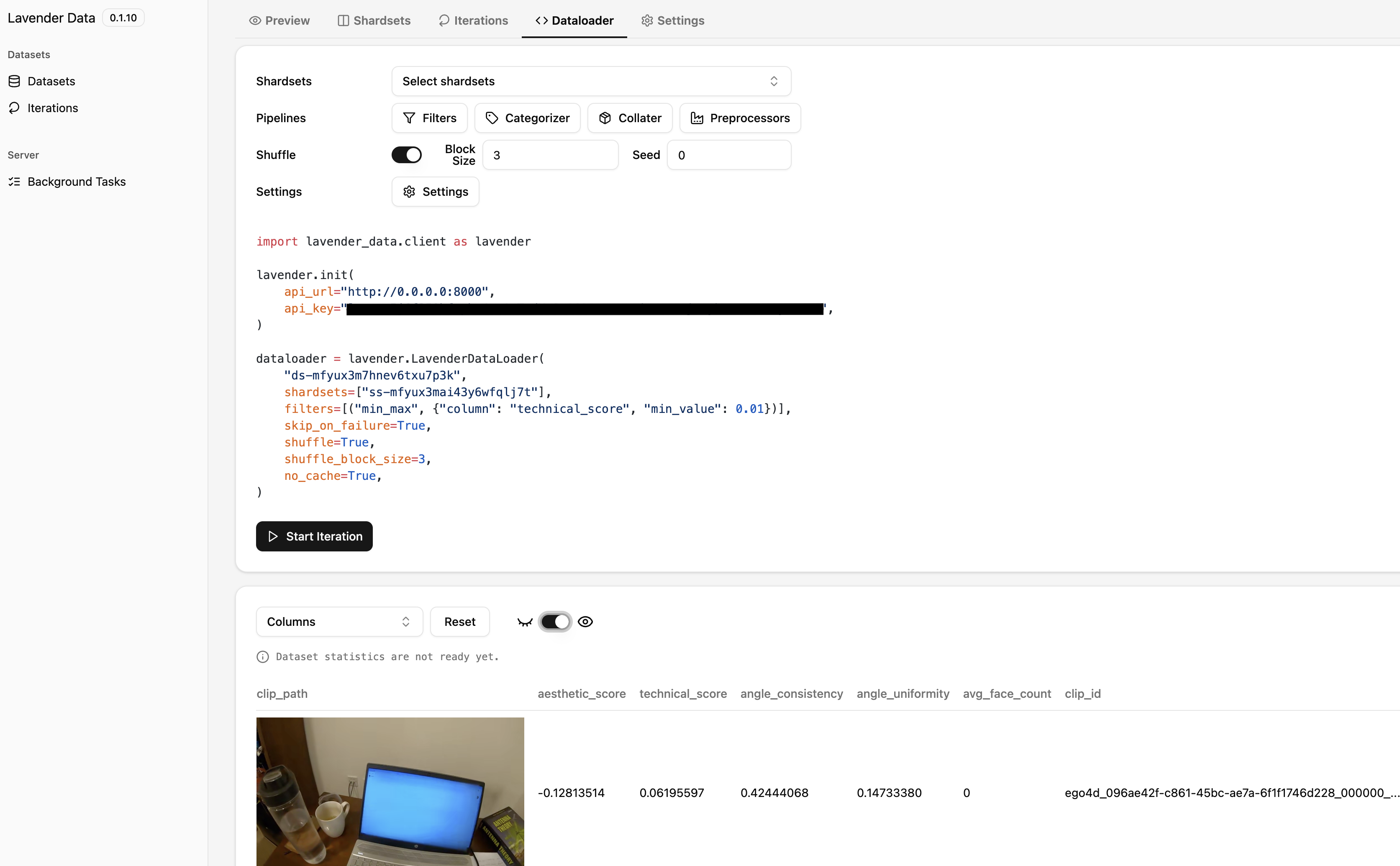}
\caption{Lavender Data web interface showing data loading simulation. The system ensures what is displayed matches exactly what the model receives during training.}
\label{fig:lavender-loader}
\end{figure}

\section{Training Methodology}
\label{sec:training}

\subsection{Architectural Choices}

In our experiments at 100M parameter scale over 10K training steps, we observed relatively small performance differences across five distinct architectural variants. Configurations including Multi-Latent Attention (MLA), window attention mechanisms, and various local structure attention patterns showed similar performance when trained with appropriately tuned learning rates. We note that these observations are specific to our experimental regime and may not generalize to larger scales or longer training horizons.

This observation led to the adoption of a vanilla transformer architecture with minimal domain-specific modifications. The primary exception is the use of three-dimensional Rotary Position Embeddings (RoPE) to encode spatiotemporal positions, as described in Section \ref{sec:rope}.

\subsubsection{Architecture Implementation Details}

Our video model is designed for high stability and avoids common failure modes associated with large-scale video diffusion pretraining. We operate entirely in the latent space, leveraging the Wan2.2-TI2V-5B VAE, which compresses raw input videos using a temporal factor of 8 and a spatial factor of 16, producing latents with 48 channels. For instance, an input of shape $(3, 81, 352, 640)$ in pixel space is converted to $(48, 21, 22, 40)$ in latent space. Before entering the transformer, latents are normalized channel-wise using precomputed VAE statistics: $z_{\text{norm}} = (z - \mu) / \sigma$, which promotes numerical stability.

For patching, we use conservative 3D patch sizes ($p_t=1$, $p_s=2$), generating $21 \times 11 \times 20 = 4620$ patches per video. Each patch is linearly projected into the transformer’s hidden dimension, ensuring consistent input scaling. To prevent mode collapse and support effective multimodal learning, we concatenate context tokens (such as text embeddings) to the video patches, so the model attends jointly over both streams. 

We ensure robust training dynamics with several architectural safeguards: 
\textbf{Adaptive Layer Normalization} \cite{huang2017arbitrary} is employed, where layer norm is modulated by timestep-dependent parameters generated via a dedicated MLP. Critically, all residual connections are gated with learned, timestep-aware parameters; a constant offset ($1/8$) is added at initialization to guarantee healthy gradient flow from the outset, preventing dead zones in gating. 

\textbf{Value Residual Connections} \cite{zhou2024value} are introduced within the attention module, governed by mixing coefficients $\lambda_1$ and $\lambda_2$ (both initialized to 0.5): $V_{\text{out}} = \lambda_1 V_{\text{prev}} + \lambda_2 V_{\text{current}}$. This mechanism ensures direct signal propagation through deep stacks, substantially reducing vanishing gradient risks and maintaining effective credit assignment even late in training.

Across all components, parameterization and normalization schemes are chosen for maximal training stability, ensuring the model remains stable across all scales and datasets.

\subsection{Three-Dimensional Rotary Position Embeddings}
\label{sec:rope}

Standard approaches apply separate one-dimensional RoPE \citep{rope} encodings to each spatiotemporal dimension independently, leaving many embedding dimensions as identity.

\begin{figure}[H]
\centering
\includegraphics[width=0.6\textwidth]{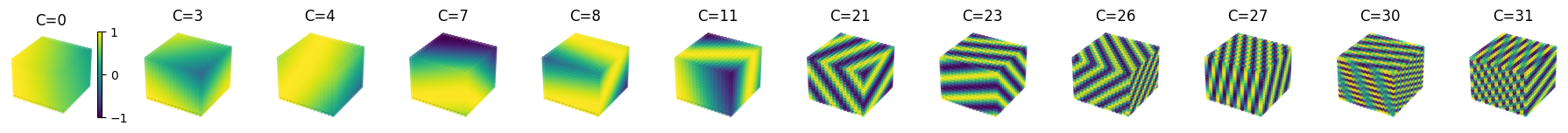}
\caption{Three-dimensional rotation axes for RoPE in a single transformer layer. Each point is the axis direction for one head--frequency pair, showing uniform coverage of the sphere via $\phi$-sequence sampling.}
\label{fig:3drope}
\end{figure}

Following ND-RoPE \citep{xiong2025ndrope}, we instead assign each frequency band a random three-dimensional unit vector $\mathbf{d} \in \mathbb{R}^3$ (sampled via low-discrepancy $\phi$-sequences for uniform coverage) and compute the rotation angle as $\theta = \omega \cdot \langle \mathbf{d}, \mathbf{p} \rangle$ where $\mathbf{p} = (t, h, w)$ is the normalized position. Each layer uses a different seed for its axes, so layers encode position through distinct geometric transformations. Frequencies are log-spaced between $\omega_{\min} = 0.2$ and $\omega_{\max} = 50$, with 10\% set to zero (partial RoPE \citep{partialrope}) to leave capacity for content-based attention. Figure \ref{fig:3drope} visualizes the resulting axis distribution.

\subsection{Hypersphere-Constrained Optimization}

Inspired by recent work on diffusion model training dynamics \citep{edm2}, we constrain weight matrix rows to unit scale, parameterizing weights on the unit sphere $\mathcal{S}^{d-1}$. This geometric constraint provides several benefits over standard weight decay: it offers a clear and explicit regularization mechanism, ensures all parameter rows have identical norm enabling straightforward analysis of training dynamics, and removes one degree of freedom per row while preserving sufficient model capacity due to inherent redundancy in neural network parameterization.

A key distinction from \citet{edm2} is that we maintain weights on the hypersphere \emph{throughout} training rather than applying normalization ad-hoc during forward passes. This approach naturally frames our optimization as \emph{Riemannian gradient descent} on the sphere manifold \citep{riemannian_optim, stiefel}. Riemannian optimization ensures that updates respect the manifold geometry by: (1) computing gradients in the \textbf{tangent space} $T_w\mathcal{S}^{d-1}$ at the current iterate, and (2) mapping updates back to the manifold via a \textbf{retraction}.

For each weight matrix $W$, we constrain row $i$ to satisfy $\|w_i\|_2 = 1$. The tangent space at $w_i$ consists of all vectors orthogonal to $w_i$. Given the Euclidean gradient $g_i \in \mathbb{R}^d$, we obtain the Riemannian gradient via the orthogonal projector $\text{Proj}_{w_i}: \mathbb{R}^d \to T_{w_i}\mathcal{S}^{d-1}$:
\begin{equation}
g_{\text{tangent}} = \text{Proj}_{w_i}(g_i) = (I - w_i w_i^\top) g_i = g_i - (g_i \cdot w_i) w_i
\end{equation}
This \textbf{tangent space projection} removes the radial component of the gradient, ensuring updates remain tangent to the sphere.

We apply AdamW with first and second moment statistics computed using the projected gradient $g_{\text{tangent}}$ rather than raw gradients. The complete update rule for hypersphere-constrained parameters is:

\begin{align}
g_{\text{tangent}} &= (I - w_t w_t^\top) g_t & \text{(\textbf{tangent space projection})} \\
m_t &= \beta_1 m_{t-1} + (1-\beta_1) g_{\text{tangent}} \\
v_t &= \beta_2 v_{t-1} + (1-\beta_2) g_{\text{tangent}}^2 \\
\hat{m}_t &= m_t / (1 - \beta_1^t), \quad \hat{v}_t = v_t / (1 - \beta_2^t) \\
\tilde{w}_{t+1} &= w_t - \eta \frac{\hat{m}_t}{\sqrt{\hat{v}_t} + \epsilon} \\
w_{t+1} &= \text{Retr}_{w_t}(\tilde{w}_{t+1} - w_t) = \frac{\tilde{w}_{t+1}}{\|\tilde{w}_{t+1}\|} & \text{(\textbf{retraction})}
\end{align}

The final normalization step implements the \textbf{retraction} $\text{Retr}_w: T_w\mathcal{S}^{d-1} \to \mathcal{S}^{d-1}$, mapping the updated point back onto the sphere. This retraction approximates the exponential map on $\mathcal{S}^{d-1}$ and is exact for infinitesimal steps. The retraction operation requires numerical care in reduced precision arithmetic, necessitating full-precision optimizer state and gradient scaling.

We distinguish between norm-preserving parameters (two-dimensional weight matrices in linear layers) and non-preserving parameters (layer normalization parameters, scalars, and biases). Our implementation automatically detects parameter types: any 2D parameter belonging to a Linear layer is classified as norm-preserving, except for specific layers including final projection (\texttt{final\_proj}), modulation layers (\texttt{modulation.2}), and depthwise convolution weights (\texttt{dconv.weight}), which use standard updates without constraints as these parameters benefit from unconstrained scale adaptation.

The row-wise nature of hypersphere normalization proves particularly convenient for distributed training. FSDP2 naturally shards parameters along rows, meaning each device independently optimizes its assigned rows without requiring inter-device communication for constraint enforcement. The tangent projection and re-normalization operations are purely local to each row. We employ FSDP2 with mixed precision policy using bfloat16 for parameters and forward activations while maintaining float32 for gradient reduction. Each transformer block is independently wrapped with FSDP, with resharding disabled on the final block to reduce communication overhead. This block-level sharding strategy enables training models with up to 30 billion parameters across 224 GPUs (28 nodes $\times$ 8 H100 GPUs) while maintaining high arithmetic intensity and minimizing communication bottlenecks.

\subsection{Maximal Update Parameterization}

A fundamental challenge in scaling neural networks is the sensitivity of optimal learning rates to model size and training duration. Standard parameterization requires extensive hyperparameter search at each scale, consuming significant computational resources.

\begin{figure}[H]
\centering
\includegraphics[width=0.7\textwidth]{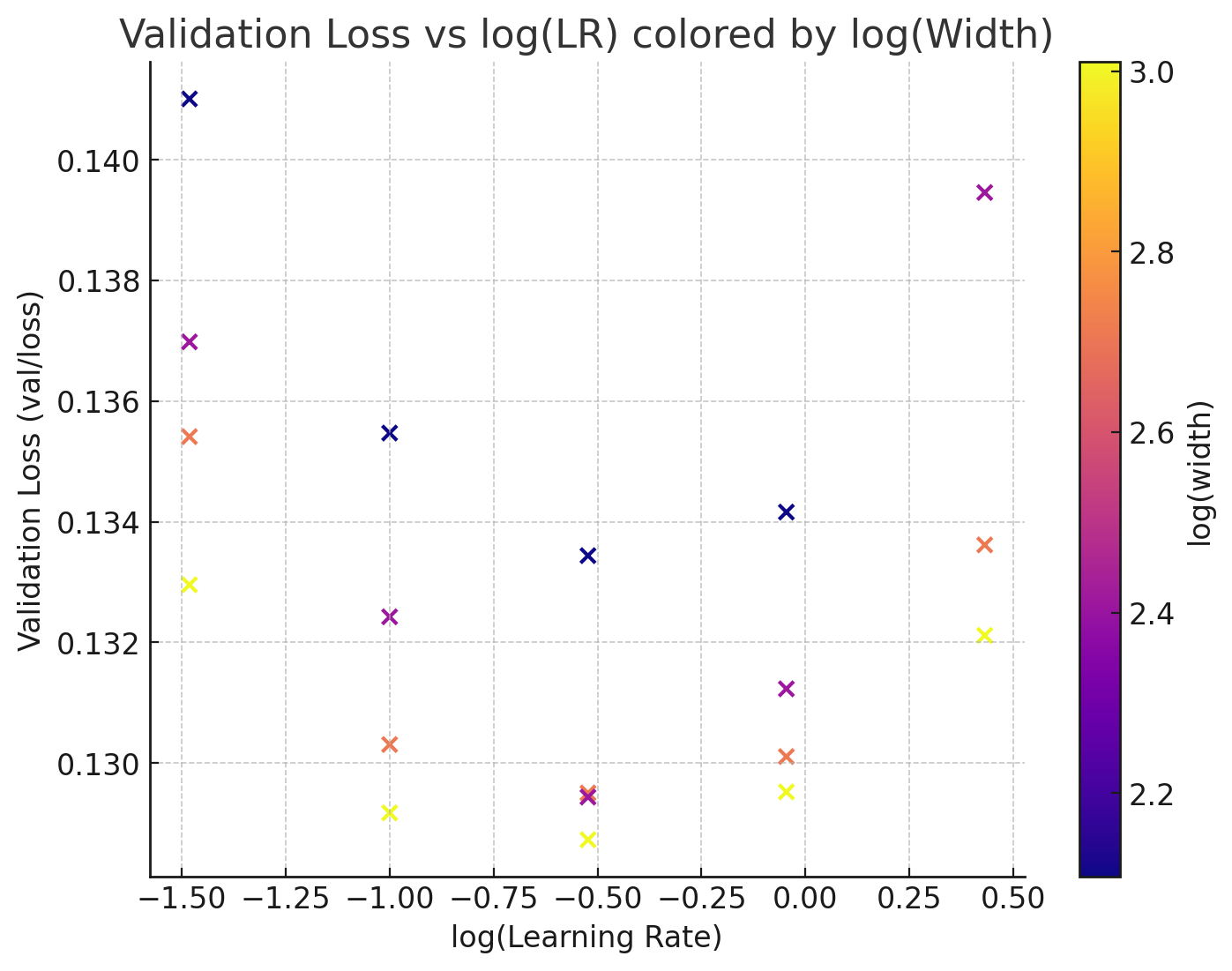}
\caption{Empirical validation that $\mu$P hyperparameter transfer works under hypersphere-constrained Riemannian optimization. Optimal learning rates transfer across model scales (30M to 1B parameters), demonstrating compatibility between $\mu$P and our geometric constraints.}
\label{fig:mup_adam}
\end{figure}

Maximal Update Parameterization ($\mu$P) \citep{mup} addresses this challenge by providing scaling rules that preserve optimal learning rates as model dimensions change. To our knowledge, we are the first to combine $\mu$P with hypersphere-constrained Riemannian optimization. This combination is theoretically motivated: hypersphere constraints naturally align with $\mu$P initialization, as constraining each row to unit norm yields $\text{std}(W_{ij}) \sim 1/\sqrt{\text{fan\_in}}$, which matches the $\mu$P scaling for weight matrices. The geometric constraint implicitly enforces the correct initialization scale, and we empirically verify that $\mu$P hyperparameter transfer remains effective under this setting (Figure \ref{fig:mup_adam}).

For our architecture, $\mu$P is implemented through a rule-based system that assigns per-parameter learning rates based on parameter names and width. Learning rate rules assign $\eta_w = \eta_{\text{base}} / \text{width}$ for most weight matrices, with intermediate transformer layers using reduced rates of $0.1 \cdot \eta_{\text{base}} / \text{width}$ for improved stability. Biases, scalar parameters, and positional embeddings use width-independent rates of $0.01 \cdot \eta_{\text{base}}$. Modulation output layers and positional embeddings initialize to zero for stable early training. The base learning rate $\eta_{\text{base}} = 0.01$ incorporates warmup, constant phase, and cooldown starting at 98 percent of total training steps. These layerwise learning rate schedules reflect the empirical observation that different architectural components benefit from different learning rates, with input-output layers requiring higher rates than intermediate processing layers.

In our experiments, $\mu$P enabled transfer of hyperparameters found at 30 million parameter scale to 1 billion parameter models, and from 1,000 training steps to 100,000 steps with minimal adjustment. This reduced the cost of hyperparameter optimization in our setting.

\subsection{Scaling Laws for Batch Size and Training Duration}

Beyond model size, training configuration choices such as batch size and training horizon significantly impact optimal learning rates. Through extensive ablation studies, we identified predictable scaling relationships.

In our experiments, we observed that optimal learning rate scaled approximately as $\sqrt{B}$ where $B$ denotes batch size, and decreased with training duration approximately as $1/\sqrt{T}$ where $T$ represents the total number of training steps. We note that these are empirical observations from our specific setting (video diffusion, our architecture and data) and may not generalize broadly. These relationships suggest a practical heuristic for our setting: when scaling computational resources by a factor of $n$, increase batch size by $\sqrt{n}$ and extend training duration by $\sqrt{n}$. Under this scaling strategy, the optimal learning rate remained approximately constant in our experiments, simplifying hyperparameter transfer across different computational budgets.

\subsection{Monitoring Training Dynamics}

Stable training at scale requires comprehensive monitoring of training dynamics beyond standard loss curves. At each validation checkpoint, we record gradient magnitudes, weight matrix norms, and weight update magnitudes for every layer.

$\mu$P provides theoretical predictions for the evolution of these statistics, forming what we term the ``$\mu$P band'': a predictable range within which well-behaved parameters should remain throughout training. Departures from this band serve as early warning signals of training instabilities. Notably, modulation parameters (lambda parameters in diffusion models) represent expected exceptions to the $\mu$P band, as these parameters adaptively scale intermediate representations.

\begin{figure}[H]
\centering
\begin{subfigure}[b]{0.48\textwidth}
\centering
\includegraphics[width=\textwidth]{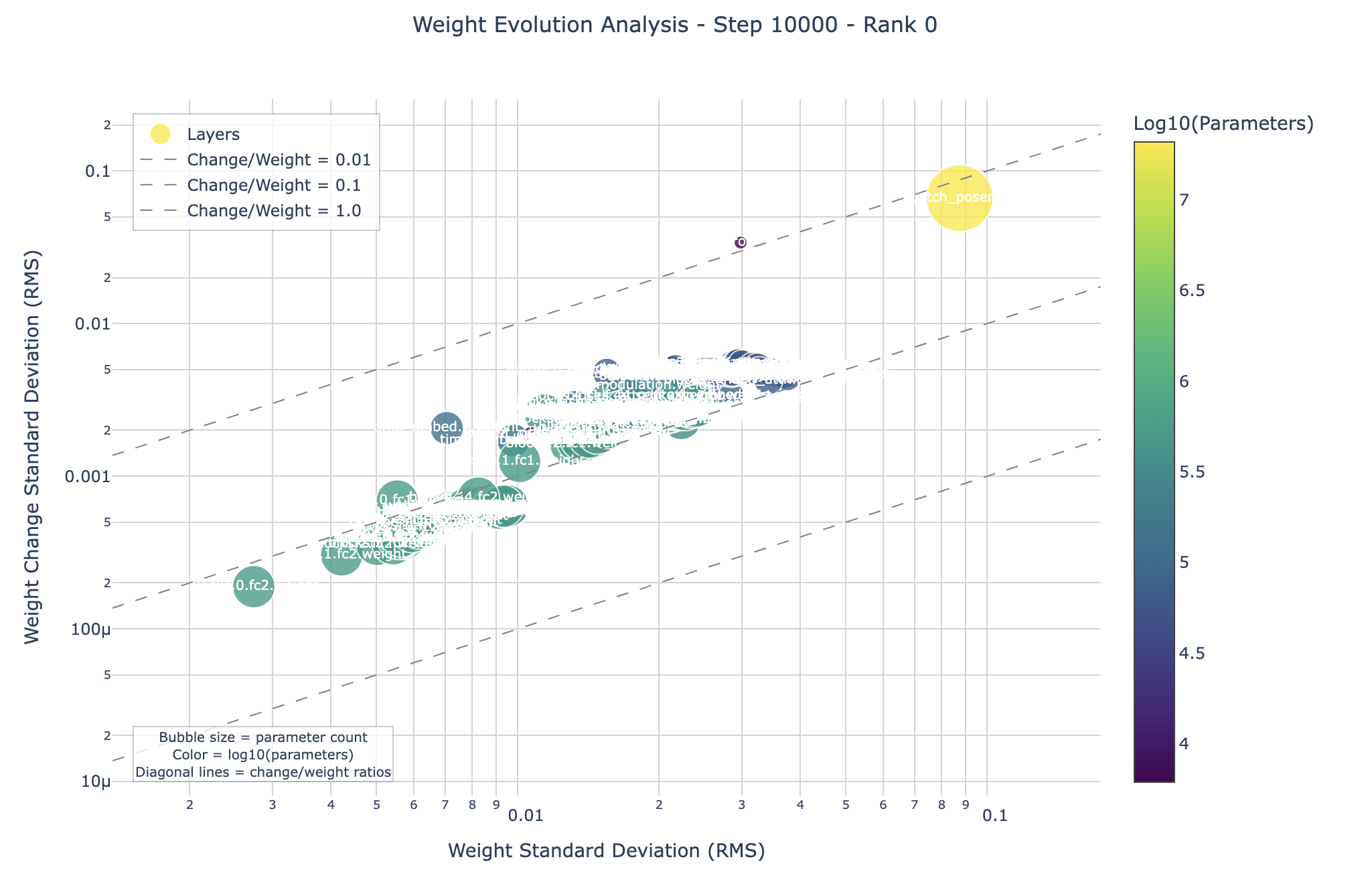}
\caption{Healthy training dynamics}
\label{fig:evolution_good}
\end{subfigure}
\hfill
\begin{subfigure}[b]{0.48\textwidth}
\centering
\includegraphics[width=\textwidth]{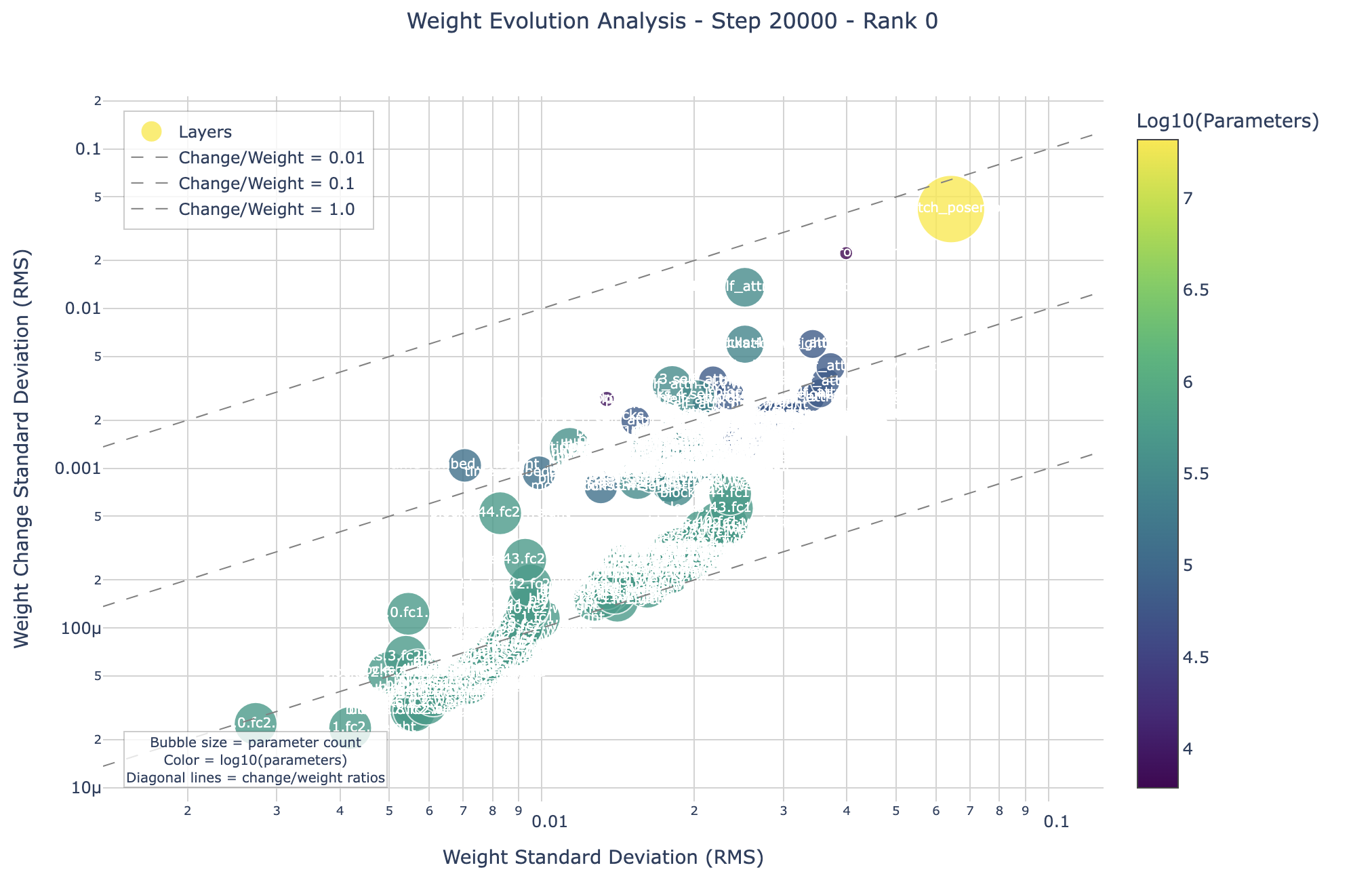}
\caption{Unhealthy training dynamics}
\label{fig:evolution_bad}
\end{subfigure}
\caption{Comparison of training dynamics. (a) shows weights remaining within the $\mu$P band with smooth evolution. (b) shows weights escaping the predicted range, indicating training instability.}
\label{fig:training_dynamics}
\end{figure}

Figures \ref{fig:evolution_good} and \ref{fig:evolution_bad} contrast healthy and unhealthy training dynamics. In well-behaved training runs, parameter norms evolve smoothly within the predicted range. Erratic behavior or escape from the $\mu$P band indicates potential issues requiring intervention.

\subsection{Hyperparameter Optimization with Bayesian Methods}

For fine-grained hyperparameter tuning, particularly layer-specific learning rate multipliers, we employ HEBO (Heteroscedastic Evolutionary Bayesian Optimization) \citep{hebo}. Bayesian optimization provides sample-efficient exploration of hyperparameter spaces, crucial when evaluation costs are high.

\begin{figure}[H]
\centering
\includegraphics[width=0.7\textwidth]{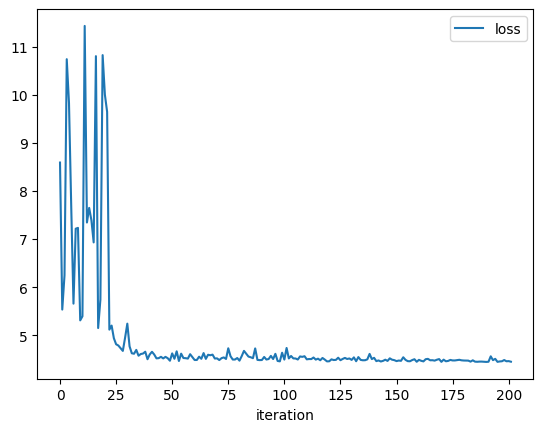}
\caption{HEBO optimization convergence for layer-specific learning rates. Multiple independent runs (shown in different colors) converge to similar performance levels within 50 iterations, indicating redundancy in the sensitivity to these hyperparameters.}
\label{fig:hebo}
\end{figure}

Figure \ref{fig:hebo} shows convergence behavior of HEBO across multiple optimization runs. Notably, the optimizer converges to near-optimal configurations within approximately 50 iterations, after which performance plateaus. This rapid convergence suggests substantial redundancy in the optimizer's sensitivity to layer-specific learning rates, implying that a wide range of layer-wise learning rate configurations yield similar performance.

\subsection{Post-hoc Exponential Moving Average}

\begin{figure}[H]
\centering
\includegraphics[width=0.7\textwidth]{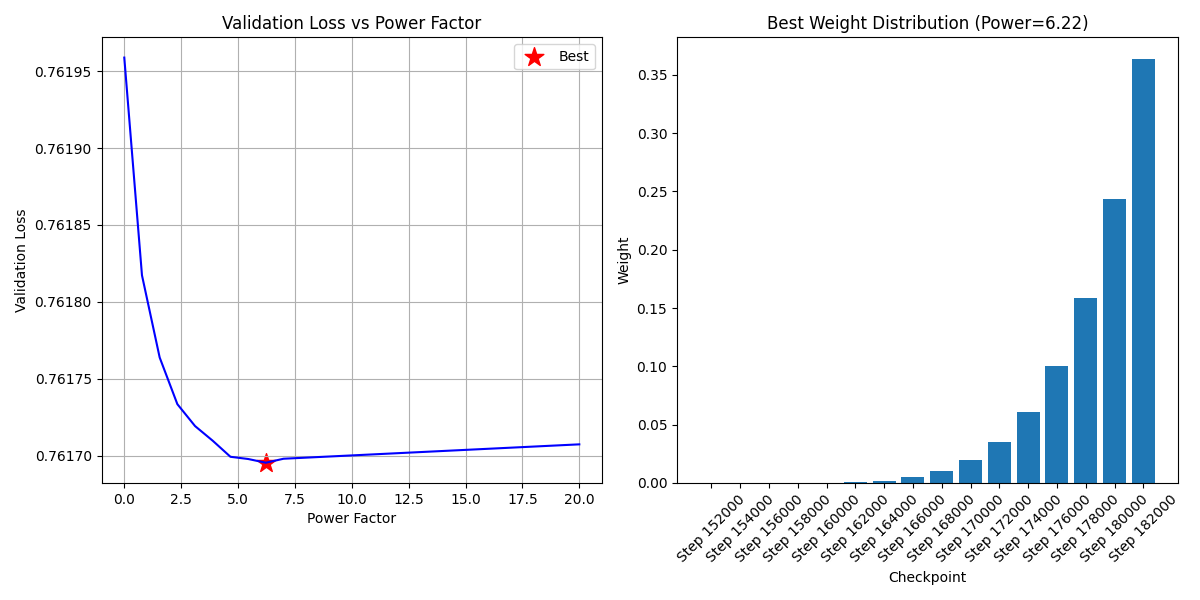}
\caption{Weight distribution for post-hoc exponential moving average using power schedule. Later checkpoints receive substantially higher weight, reflecting their superior quality. The optimal $\alpha$ value is selected based on validation loss.}
\label{fig:ema}
\end{figure}

Rather than maintaining an exponential moving average (EMA) of parameters online during training, we employ post-hoc averaging of checkpoints with a power-law schedule following \citet{edm2}. Specifically, the weight assigned to checkpoint at step $t$ is:

\begin{equation}
\beta(t) = \left(1 - \frac{1}{t+1}\right)^{1+\alpha}
\end{equation}
where $\alpha$ controls the emphasis on later checkpoints. We use $\alpha = 6.22$, providing strong preference for later training stages where model quality improves more rapidly. We select this weight assignment based on validation loss, as demonstrated in Figure \ref{fig:ema}.

\subsection{Inference Aware  Architectural Design}

Video models generate approximately 100,000 tokens for typical outputs, making attention operations extremely expensive at inference time. To mitigate this computational burden, we adopt a parallel MLP architecture where attention and feed-forward computations occur in parallel branches rather than sequentially.

Our implementation projects the input through a unified linear layer that produces query, key, value, and MLP hidden states simultaneously:
\begin{equation}
[Q, K, V, H_1, H_2] = \text{Linear}_{7d}(\text{RMSNorm}(x))
\end{equation}
where the dimensions are split as $3d$ for attention (Q, K, V) and $4d$ for MLP computation. The MLP uses a gated activation:
\begin{equation}
\text{MLP}(x) = \text{Linear}_2(\text{GELU}(H_2) \odot H_1)
\end{equation}

The complete transformer block operates as:
\begin{align}
\tilde{x} &= \text{RMSNorm}(x) \cdot (1 + s_t) + b_t \\
[Q, K, V, H_1, H_2] &= \text{Split}(\text{Linear}(\tilde{x})) \\
A &= \text{Attention}(\text{RoPE}(Q), \text{RoPE}(K), V) \\
M &= \text{Linear}_2(\text{GELU}(H_2) \odot H_1) \\
y &= x + \text{Linear}_{\text{proj}}([A, M]) \cdot (g_t + \frac{1}{\sqrt{L}})
\end{align}

where $s_t$, $b_t$, and $g_t$ are time-dependent modulation parameters computed from timestep embeddings, and $L$ denotes the total number of transformer layers.

The constant offset $1/\sqrt{L}$ in the gating term prevents the block from being mathematically inert at initialization. We initialize $\text{Linear}_{\text{proj}}$ with zero weights and $g_t = 0$, which yields an identity map in the forward pass: $y = x$. However, without the offset, the gradient with respect to the projection weights would be zero:
\begin{equation}
\frac{\partial \mathcal{L}}{\partial W_{\text{proj}}} = \frac{\partial \mathcal{L}}{\partial y} \cdot (g_t + 0) \cdot [A, M]^\top = 0
\end{equation}
This makes the entire block untrainable for non-negligible number of steps: no gradient flows to the projection, attention, or MLP components. The $1/\sqrt{L}$ term seeds gradient flow while maintaining depth-aware scaling: across $L$ layers, residual contributions sum as $\sum_{\ell=1}^{L} \frac{1}{\sqrt{L}} \Delta_\ell \sim \mathcal{O}(1)$, keeping activations and gradients stable regardless of depth. As training progresses, $g_t$ learns to modulate the residual magnitude, but early optimization is bootstrapped by this fixed path. This follows the same philosophy as DeepNorm and residual scaling techniques, adapted to gated zero-initialized blocks.

This design enables overlap of attention computation (which may involve expensive all-to-all communication in distributed settings) with MLP computation (which is typically compute-bound). The parallel structure is conceptually similar to recent work on parallel attention mechanisms, though our formulation uses unified projection for simplicity. In our benchmarks, this architecture reduced inference latency by approximately 20 percent compared to a sequential baseline, while we observed no meaningful difference in training dynamics.

\section{Experiments and Analysis}
\label{sec:experiments}

\subsection{Architectural Ablations}

We evaluated five distinct architectural variants at 100M parameter scale over 10K training steps each. The variants included: standard transformer with 3D RoPE, transformer with Multi-Latent Attention (MLA), window attention with various window sizes, transformers with different local attention patterns, and attention with differential mechanisms \citep{diffattention}.

When trained with appropriately tuned learning rates, all variants achieved similar final performance in our experiments, with differences smaller than run-to-run variance. While we cannot rule out that architectural differences would emerge at larger scales, longer training, or different evaluation metrics, this observation informed our decision to prioritize optimization and data quality over architectural exploration for the remainder of this project.

\begin{figure}[H]
\centering
\includegraphics[width=0.8\textwidth]{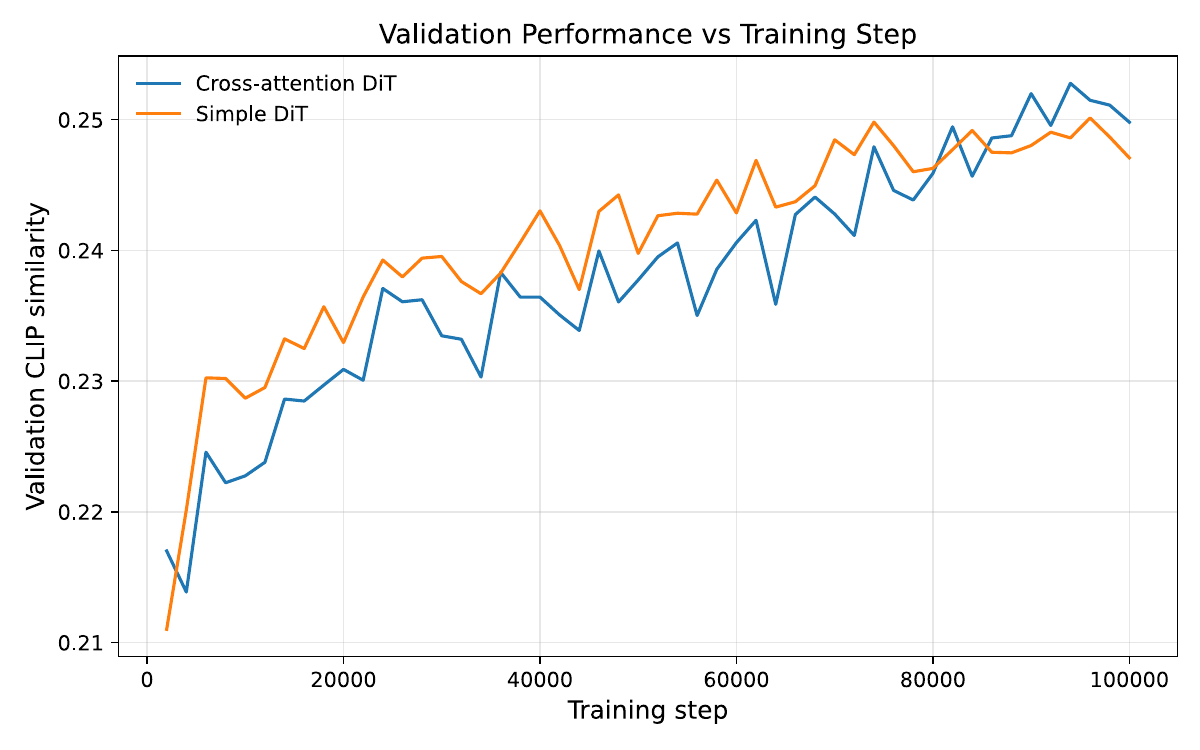}
\caption{Comparison of validation CLIP similarity between Cross-attention DiT and Simple DiT architectures over training. Both architectures achieve similar final performance, demonstrating that architectural choices have minimal impact compared to optimization and data quality.}
\label{fig:arch_comparison}
\end{figure}

Figure \ref{fig:arch_comparison} illustrates this finding, comparing validation CLIP similarity between a Cross-attention DiT architecture and a simpler DiT variant over the course of training. Despite their structural differences, both architectures converge to nearly identical performance, with overlapping trajectories throughout training.

One notable exception emerged during scaling experiments. Differential Attention, which showed promise at small scale, exhibited training instabilities when scaled to billion-parameter models. We did not investigate the root cause extensively, instead opting for the more stable vanilla attention mechanism.

Similarly, QK normalization, initially included in our architecture, proved unnecessary for training stability. Following the principle of preferring simpler designs, we removed this component without observing any degradation in training dynamics or final performance.

\subsection{Scaling Experiments}

To validate $\mu$P's hyperparameter transfer properties, we conducted experiments across three model scales (30M, 100M, and 1B parameters) and three training durations (1K, 10K, and 100K steps). For each configuration, we performed learning rate sweeps and recorded final validation loss.

The results were consistent with $\mu$P's predictions in our setting. Learning rates optimized at 30M parameters transferred to 1B parameters with minimal adjustment required. Similarly, learning rates found effective at 1K steps remained effective at 100K steps when adjusted according to our empirically observed $\sqrt{T}$ scaling relationship. We note that these scaling relationships are empirical observations from our experiments and may require validation in other settings.

Combined with the batch size scaling rule ($\text{LR} \propto \sqrt{B}$), these relationships enable the following workflow: identify optimal hyperparameters using a small model trained for a short duration, apply $\mu$P scaling rules to obtain learning rates for the production model size, and adjust for planned batch size and training duration using the empirical scaling laws.

This procedure reduced our hyperparameter search costs substantially compared to what direct optimization at production scale would have required.

\subsection{Training Stability Analysis}

Throughout training, we monitored the $\mu$P band for all parameter groups. In successful runs, weight norms evolved smoothly within predicted ranges across all layers except modulation parameters. Several training runs that initially appeared healthy based on loss curves alone showed subtle deviations in weight norm dynamics, which later manifested as training instabilities.

These observations validate the importance of comprehensive monitoring beyond scalar metrics. The geometric interpretation provided by hypersphere constraints further aids in diagnosing issues: row norms that drift away from unity indicate either numerical precision problems in the projection operation or inappropriate learning rates for those specific parameters.

\subsection{Dataset Quality Impact}

To assess the impact of our multi-stage filtering pipeline, we trained models on progressively filtered versions of the dataset. Starting from raw segmented clips, we incrementally added visual filters (color, thumbnails), motion filters (optical flow, foreground/background), and quality filters (DOVER scores).

In our experiments, each filtering stage improved downstream model performance as measured by validation loss and CLIP similarity. We observed that motion-based filtering and DOVER scores appeared to contribute meaningfully, though we did not conduct controlled ablations isolating each component. Removing clips with insufficient motion (likely slideshows or static content) appeared beneficial, consistent with the intuition that dynamic content aids learning of temporal dynamics.

The deduplication process removed approximately 20\% of clips within each semantic bucket. In our experiments, models trained on the deduplicated dataset showed improved validation metrics without requiring longer training, though we did not conduct extensive ablations to isolate this effect. This is consistent with the hypothesis that near-duplicates contribute diminishing returns to learning.

\subsection{Captioning Quality}

The hierarchical captioning strategy proved essential for dataset balancing. Without ultra-short captions for bucketing, the dataset exhibited severe imbalance toward certain categories, particularly videos featuring people speaking directly to the camera.

We evaluated caption quality using a holdout set manually annotated with ground truth descriptions. The fine-tuned Qwen 2.5 VL model achieved high agreement with human annotations for short and ultra-short captions, while detailed captions showed greater variance in style and content coverage. For our purposes, consistency in ultra-short captions proved more important than detailed caption accuracy, as these serve primarily for deduplication rather than as training signals.

\section{Benchmark Evaluation}
\label{sec:benchmark}

We evaluate our model against video generation systems of comparable scale using VBench 1.0 \citep{huang2024vbench} and VBench 2.0 \citep{zheng2025vbench} benchmarks. These benchmarks provide comprehensive evaluation across multiple dimensions including visual quality, temporal consistency, motion dynamics, and semantic alignment.

\begin{table}[H]
\centering
\caption{VBench 2.0 evaluation results across five quality dimensions. Summer-22B is compared against Wan 2.2 variants (5B and A14B, open-source models of similar or larger scale) and Veo3 Fast (a proprietary system included as an upper reference). Higher is better for all metrics. Our model is competitive on commonsense and physics but shows gaps in creativity and controllability, likely reflecting limited prompt diversity during training.}
\label{tab:vbench2}
\begin{tabular}{lccccc}
\toprule
Model & Creativity & Commonsense & Controllability & Human Fidelity & Physics \\
\midrule
Summer-22B (Ours) & 0.387 & 0.622 & 0.311 & 0.745 & 0.629 \\
Wan 2.2-5B & 0.503 & 0.581 & 0.324 & 0.803 & 0.662 \\
Wan 2.2-A14B & 0.516 & 0.595 & 0.454 & 0.793 & 0.690 \\
Veo3 Fast & 0.556 & 0.619 & 0.415 & 0.822 & 0.679 \\
\bottomrule
\end{tabular}
\end{table}

\begin{table}[H]
\centering
\caption{VBench 2.0 total scores (weighted average across all dimensions). Summer-22B achieves 0.539, within 0.036 of the open-source Wan 2.2-5B baseline trained on substantially more data, demonstrating the effectiveness of our data-efficient training methodology.}
\label{tab:vbench2_total}
\begin{tabular}{lc}
\toprule
Model & Total Score \\
\midrule
Summer-22B (Ours) & 0.539 \\
Wan 2.2-5B & 0.575 \\
Wan 2.2-A14B & 0.610 \\
Veo3 Fast & 0.618 \\
\bottomrule
\end{tabular}
\end{table}

\begin{figure}[H]
\centering
\includegraphics[width=0.49\textwidth]{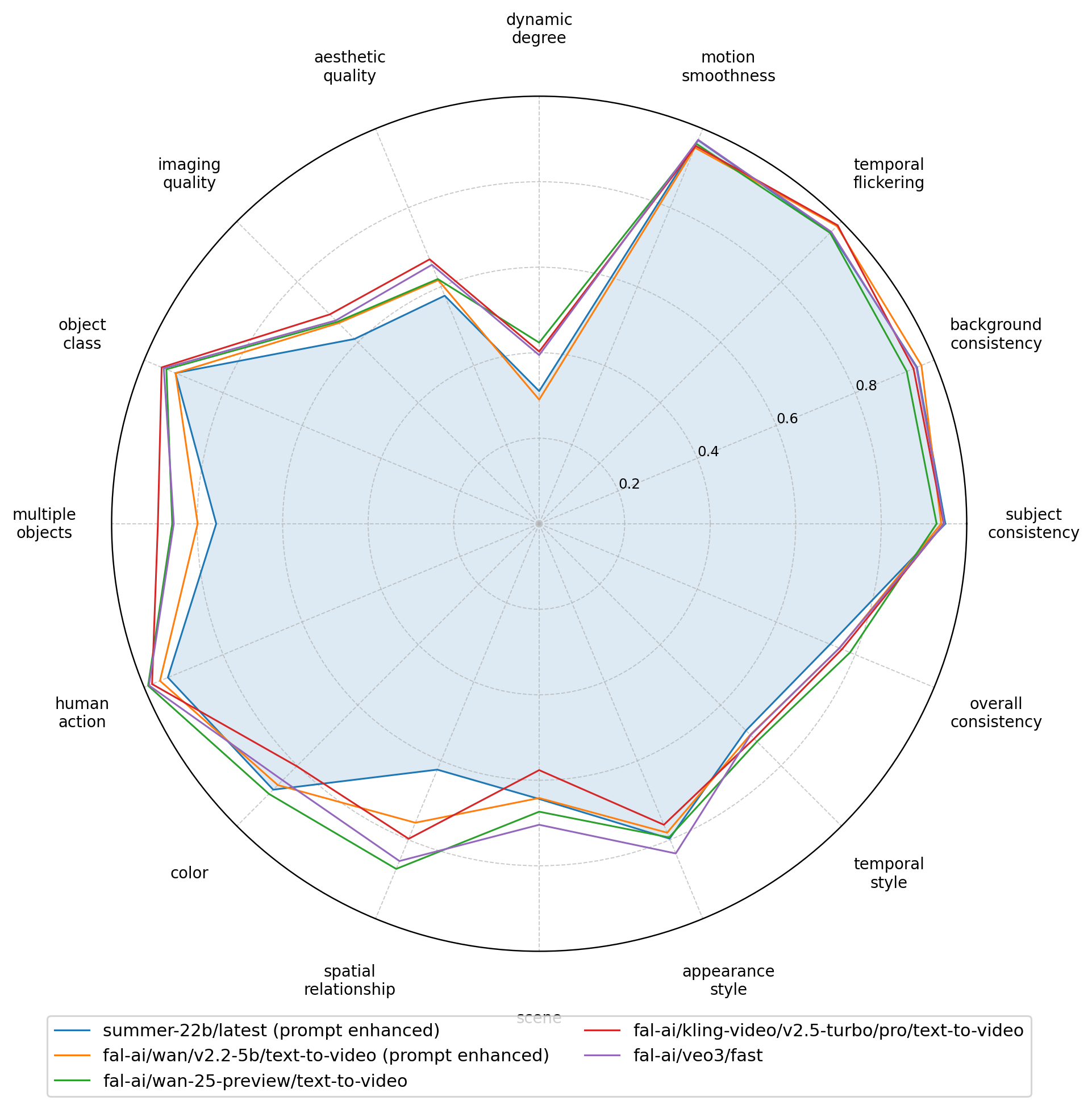}
\hfill
\includegraphics[width=0.49\textwidth]{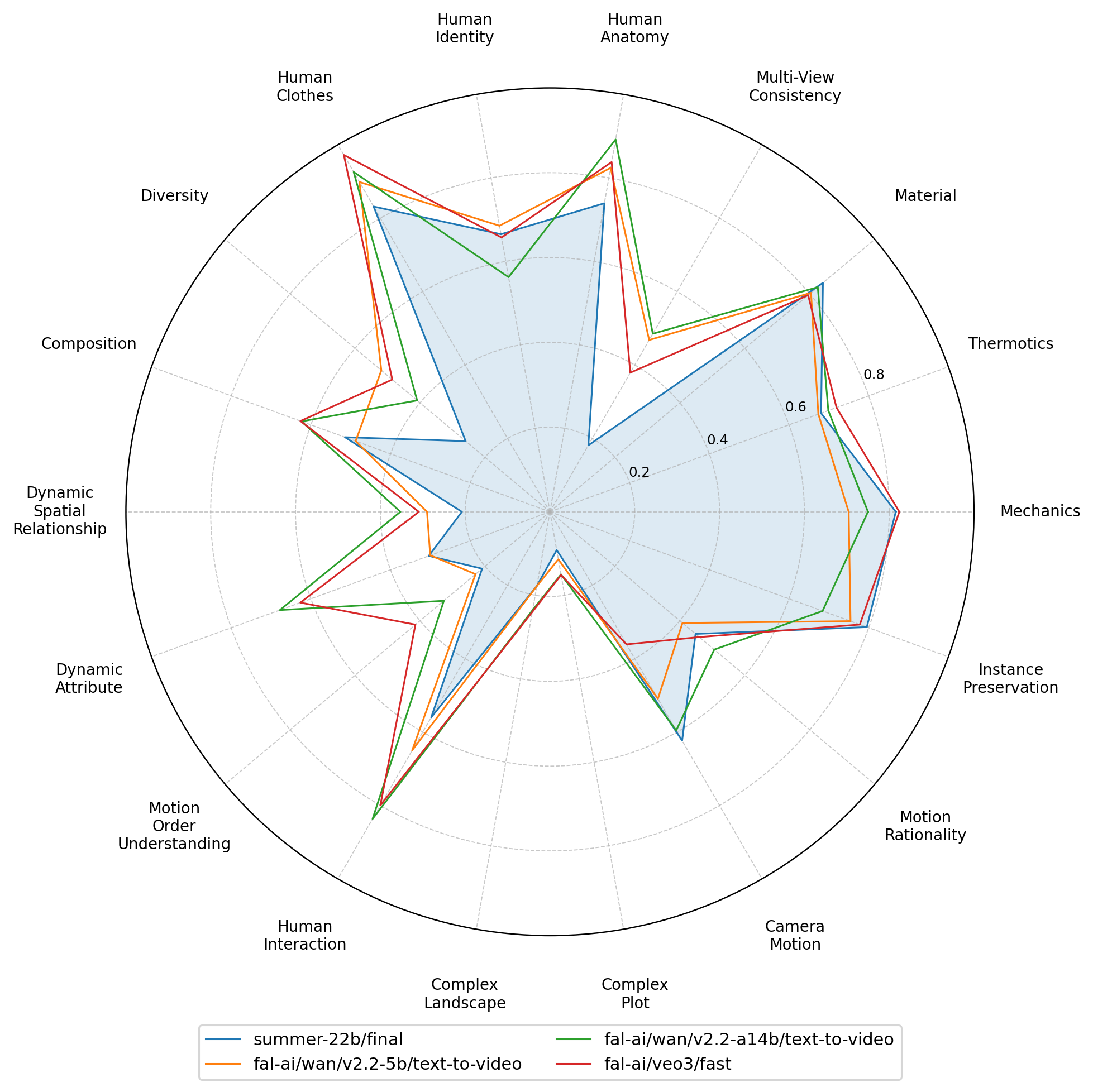}
\caption{Radar plots of per-dimension VBench scores. \textbf{Left:} VBench 1.0 evaluates 16 quality dimensions including temporal consistency, motion smoothness, and semantic alignment. \textbf{Right:} VBench 2.0 evaluates 20 advanced dimensions covering creativity, physics, and human fidelity. Summer-22B (blue) is compared against Wan 2.2-5B, Wan 2.2-A14B, and Veo3 Fast. Our model is competitive on low-level quality metrics (flickering, smoothness) but lags on high-level semantic dimensions.}
\label{fig:vbench}
\end{figure}

Figure \ref{fig:vbench} presents detailed radar plots comparing performance across individual VBench dimensions. On VBench 1.0, our model shows reasonable temporal consistency (flickering, background/subject consistency) and motion smoothness, but lags behind in semantic dimensions such as spatial relationships and appearance style. The VBench 2.0 evaluation shows gaps in creativity, controllability, and complex scene understanding. We include these benchmarks for transparency, comparing against publicly available systems of similar scale (Wan 2.2 variants) as well as Veo3 for reference.

\section{Conclusion}
\label{sec:conclusion}

We have presented Summer-22B, a video foundation model trained from scratch, along with the complete methodology spanning dataset engineering, optimization, and evaluation. Our work demonstrates that with careful data curation, principled parameterization through $\mu$P, and Riemannian optimization on the sphere manifold, it is feasible to train competitive video diffusion models at scale.

The primary observations from our experience are as follows:
\begin{itemize}
\item \textbf{Dataset engineering dominates effort.} Building robust preprocessing pipelines---shot detection, multi-stage filtering, hierarchical captioning, and deduplication---consumed the majority of development time. The Lavender Data system proved essential for managing this complexity.
\item \textbf{Architectural variants show smaller differences than expected.} Within our tested regime (up to 1B parameters, 100K steps), vanilla transformers with 3D RoPE performed comparably to more complex alternatives, informing our decision to prioritize optimization and data quality.
\item \textbf{$\mu$P transfers under Hypersphere constraints.} We demonstrate, to our knowledge for the first time, that $\mu$P hyperparameter transfer remains effective when combined with hypersphere-constrained optimization. The geometric constraint naturally aligns with $\mu$P initialization ($\text{std} \sim 1/\sqrt{\text{fan\_in}}$), simplifying the training recipe.
\item \textbf{Monitoring beyond loss is essential.} Tracking parameter dynamics within the $\mu$P band provided early warning of instabilities that loss curves alone would miss. The geometric interpretation from hypersphere constraints further aids diagnosis.
\item \textbf{Cost Effective Pretraining.} Summer-22B achieves a VBench 2.0 total score of 0.539, compared to Wan 2.2-5B (0.575) and Wan 2.2-A14B (0.610) at similar scale. The total project cost of approximately \$300K (\$150K compute, remainder on infrastructure and dataset engineering) demonstrates the accessibility of video foundation model development.
\end{itemize}

Future work will explore alternative manifold constraints beyond hyperspheres, investigate the redundancy in layer-wise learning rate sensitivity suggested by our HEBO experiments, and continue scaling using the established methodology. We plan to open-source the Lavender Data system to enable broader adoption of modern data loading practices, and to release model weights along with detailed training artifacts to facilitate reproducibility and further research in video foundation models.

\section*{Acknowledgments}

We thank Burkay Gur, Gorkem Yurtseven, and Batuhan Taskaya for the support that made this project possible, and Seungju Han for detailed feedback on an earlier draft. We also benefited from discussions with colleagues across the ML community on data processing and training methodology.

\bibliographystyle{plainnat}
\bibliography{references}

\end{document}